%% file: 0_main_cameraready.tex
\def\year{2021}\relax
\documentclass[letterpaper]{article} 
\usepackage{aaai21}  
\usepackage{times}  
\usepackage{helvet} 
\usepackage{courier}  
\usepackage[hyphens]{url}  
\usepackage{graphicx} 
\usepackage{natbib}  
\usepackage{caption} 
\usepackage{amsfonts}

\usepackage{amssymb}

\usepackage{comment}
\usepackage{diagbox}
\usepackage{hyperref}
\usepackage{amsmath}
\usepackage{color}
\usepackage{xcolor}
\usepackage{textcomp}
\usepackage{gensymb}

\newcommand\todo[1]{\textcolor{red}{#1}}

\usepackage[switch]{lineno}  %
\input{0_macro}
\makeatletter
\newcommand{\printfnsymbol}[1]{%
  \textsuperscript{\@fnsymbol{#1}}%
}

\title{3D Object Recognition By Corresponding and Quantizing \\Neural 3D Scene Representations}

\author {
     Mihir Prabhudesai\thanks{Equal contribution},
    Shamit Lal\printfnsymbol{1},
    Hsiao-Yu Fish Tung,\\
    Adam W. Harley, Shubhankar Potdar, Katerina Fragkiadaki\\
}

\affiliations{
Carnegie Mellon University \\
\{mprabhud, shamitl, htung, aharley, smpotdar, katef\}@cs.cmu.edu 
}
\newcommand\blfootnote[1]{%
  \begingroup
  \renewcommand\thefootnote{}\footnote{#1}%
  \addtocounter{footnote}{-1}%
  \endgroup
}

\begin{document}
\maketitle
\blfootnote{Project page:}
\blfootnote{
\href{https://mihirp1998.github.io/project\_pages/3dq/}{https://mihirp1998.github.io/project\_pages/3dq/}}
\input{1_abstract2}

\input{2_intro_fishfinal2}
\input{3_related}
\input{4_model}

\input{5_exps_2}
\input{6_conclusion}

\clearpage
\bibliography{0_main_cameraready}

\clearpage
\input{9_suppl}

\end{document}

%% file: 0_macro.tex
\newcommand{\RR}{\mathrm{R}}

\newcommand{\K}{{\mathbb{K}}} 
\renewcommand{\S}{{\mathbb{S}}}

\newcommand{\crop}{\mathrm{crop}}
\newcommand{\Det}{\mathrm{Det}}
\newcommand{\Enc}{\mathrm{Enc}}
\newcommand{\Dec}{\mathrm{Dec}}

\newcommand{\ee}{\mathbf{e}}

\newcommand{\zido}{z_{id}^o} 
\newcommand{\zRo}{z_R^o}

\newcommand{\argmin}{\mathop{\mathrm{argmin}}}

\newcommand{\Rotate}{\mathrm{Rot}} 

\newcommand{\resize}{\mathrm{resize}}

\newcommand{\loss}{\mathcal{L}}

\newcommand{\M}{\textbf{M}} %

\renewcommand{\arraystretch}{1.2}

%% file: 1_abstract2.tex
\begin{abstract}
We propose a system that learns to detect objects and infer
their 3D poses in RGB-D images. Many existing systems
can identify objects and infer 3D poses, but they heavily rely
on human labels and 3D annotations. The challenge here is
to achieve this without relying on strong supervision signals.
To address this challenge, we propose a model that maps RGB-D images to a set of 3D visual feature maps in a differentiable fully-convolutional manner, supervised by predicting views. The 3D feature maps correspond to a featurization of the 3D world scene depicted in the images.
The object 3D feature representations are invariant to camera viewpoint changes or zooms, which means  feature matching can identify similar objects under different camera viewpoints.  
We can compare the 3D feature maps of two objects by searching alignment across scales and 3D rotations, and, as a result of the operation, we can estimate pose and scale changes without the need for 3D pose annotations. 
We cluster object feature maps into a set of 3D prototypes that represent familiar objects in  canonical scales and orientations. We then parse  images by inferring the prototype identity and 3D pose for each detected object.
We compare our method to numerous baselines that do not learn 3D feature visual representations or do not attempt to correspond features across scenes, and outperform them by a large margin in the tasks of object retrieval and object pose estimation. Thanks to the 3D nature of
the object-centric feature maps, the visual similarity cues are
invariant to 3D pose changes or small scale changes, which
gives our method an advantage over 2D and 1D methods.

\end{abstract}

%% file: 2_intro_fishfinal2.tex
\section{Introduction}\label{introduction}



The goal of this paper is detecting objects and inferring their 3D poses in RGBD images, with minimal human supervision. 
The ability to recognize objects under varying poses, sizes, lighting conditions, and camera viewpoints is fundamental for humans and other animals to track and interact with diverse objects. 
While humans and animals acquire this ability through evolution and  interacting with the world under a moving visual sensor---their eyes---, 
 most existing computer vision models are trained from labelled images,  acquired from stylized camera viewpoints \cite{maskrcnn, TulGupFouEfrMal17}. 

Recognizing familiar objects and detecting their 3D locations, poses and scales in images without 3D annotations remains elusive.
 In robotics, many works assume a closed world of predefined 3D object models, e.g., 3D object meshes, instead of discovering those from images \cite{narayanan2017deliberative}, and the fitting of the models to images is trained mostly supervised \cite{sundermeyer2018implicit,manhardt2018deep, sucar2020neural}. 
Few-shot object detection methods \cite{Koch2015SiameseNN, DBLP:journals/corr/VinyalsBLKW16, DBLP:journals/corr/SnellSZ17} use a support sample to quickly classify a query sample, but remain in 2D image space and do not infer 3D object orientation, rather object label.

Our key intuition in this work is to represent objects in terms of {\bf 3D feature representations} inferred from the input RGBD images, and  infer alignment between  two objects by explicitly rotating and scaling their representations during matching. 
While current state-of-the-art (SOTA) models for object detection and pose estimation represent an object as a feature vector or  2D feature maps \cite{rad2018feature, mehta2018single, maskrcnn}, our model represents objects as a 3D feature representation inferred from 2.5D (RGBD) input images, which can be explicitly scaled, rotated and compared in 3D. Different from  methods in robotics research that infer explicit 3D geometry of an object in terms of meshes or poinclouds from multi-view data \cite{narayanan2017deliberative, ten2018using, pinto2016supersizing} and depend heavily on a sufficient number of views, our model learns to infer the 3D object feature representation  from a single view upon self-training.

\begin{figure}
    \centering
    \includegraphics[width=0.38\textwidth]{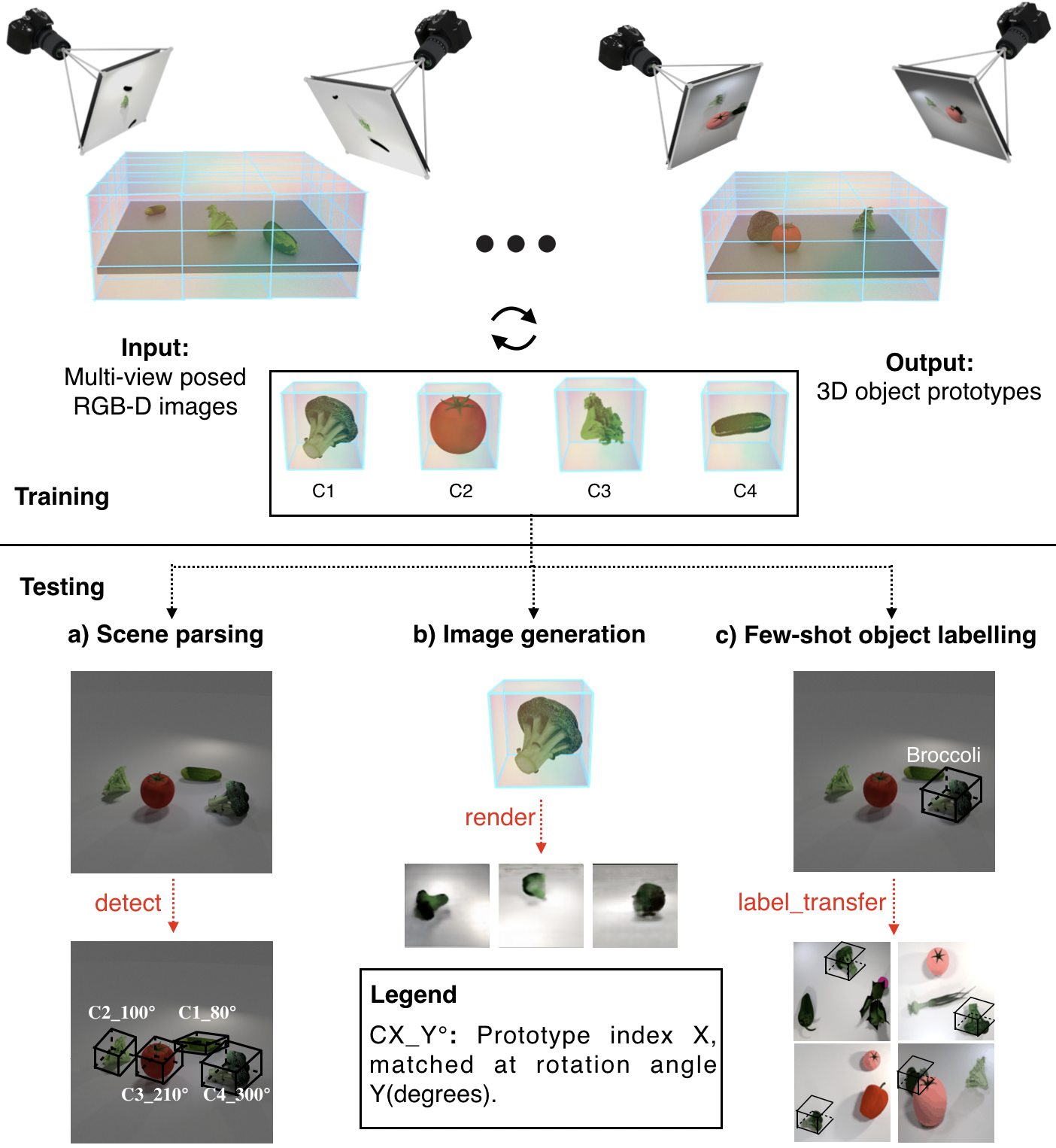}
    \caption{\textbf{Top: Model overview.} Our model takes as input posed RGB-D images of scenes, and outputs 3D prototypes of the objects. 
\textbf{Bottom: Evaluation tasks.} (a) Scene parsing: Given a new scene, we match each detected object against the prototypes using a rotation-aware check to infer its identity and pose. 
(b) Image generation: We visualize prototypes with a pre-trained 3D-to-2D image renderer. (c) Few shot object labelling: Assigning a label to a prototype automatically transfers this label to its assigned instances.}
    \label{fig:overview}
\end{figure}

We propose 3D quantized-Networks (3DQ-Nets), a model that can detect objects in 3D and that can iteratively establish accurate object correspondences without human labels or 3D annotations. We initialize its feature representations by pre-training on self-supervised view prediction task \cite{commonsense}.
 To predict views, our model maps 2.5D images
to 3D feature maps and project those to novel viewpoints to
predict 2.5D alternative views. This task is unsupervised, since to collect the data for training, we only need to put in the scene a moving agent that can freely move and observe the scene from varying viewpoint.
The inferred 3D visual feature map is view-invariant and thus unaffected by image variations caused by changes in camera viewpoint. In other words, an object will have the same representation when viewed from different camera distances and angles. 
3DQ-Nets  further improve the features through automated cross-scene correspondence mining. The step is critical for establishing more accurate correspondence between objects. 
Our model cluster objects in a pose-aware manner into several clusters of similar-looking objects. 
We call the learned cluster centers {\it prototypes}, since they correspond to aggregates of object instances across 3D poses and scales. Given a scene, our model learns to parse the scene in terms of objects associated to  prototype identities and their corresponding 3D poses (see  Figure \ref{fig:overview} (a)). The learned prototypes can be explicitly rotated, and can be rendered into images through a learned neural decoder (see Figure \ref{fig:overview} (b)). We demonstrate the usefulness of our framework in few-shot learning: our model can recognize and name objects from one or a few samples (see Figure \ref{fig:overview} (c)).  Once given a labelled instance, the model 
propagates the label to all the instances in the same cluster. 

Whether the model can infer correct correspondence from the object-centric 3D feature representation depends on the quality of two key components: the learned visual features and the 3D object detector.  
The weights of the  encoder, decoder, 3D object detector, and prototypes are optimized using a mix of end-to-end backpropagation and expectation-maximization (EM) steps, and we show 3D object detection and prototype learning  
improve over time and help one another.

We 
empirically show that the modules of our model benefit one another and are essential for  learning to recognize objects and their 3D orientation without  supervision: the 3D object detector benefits from  3D visual prototypes by discarding bounding boxes not matching to prototypes;
learning better object detection results in more accurate inference of finding object correspondences; better inferred object correspondences result in better learning of visual feature representations; and better visual feature improves clustering by inferring accurate pose-equivariant alignment of objects to prototypes.

We test our model in diverse environments including photo-realistic simulators and real world videos captured by a Kinect camera. We empirically show our model can effective learn to name new objects in a few-shot setting by propagating provided labels through the learned clusters. Our model  outperforms by a large margin numerous baselines that do not infer a 3D feature space, rather, detect and cluster objects in a 2D feature space using  CNN  feature representations pretrained on ImageNet and finetuned with the few supplied labels, or do not mine cross-scene correspondences.  We ablate each module of the proposed model and quantify its contribution in the performance of our full model. 

The main contribution of this work is 
matching objects in a  3D-aware representation space inferred from images, supervised by view prediction and automated correspondence mining, without any 3D annotations. Objects are  clustered into 3D prototypes which form then the basis for recognition: prototype identity inference and 3D pose with respect to the prototype's orientation. To the best of our knowledge, this is the first system that demonstrates that pose-aware 3D object recognition emerges without any 3D annotations in RGB-D images. Our code will be made available upon publication.

%% file: 3_related.tex
\section{Related work}

\normalfont \textbf{Self-supervised visual representation learning using pretext tasks.} 
Self-supervising visual feature representation with a variety of pretext tasks has shown to deliver useful visual representations for downstream visual recognition tasks. 
Pretext tasks that have been considered  are  predicting views of static scenes  \cite{Eslami1204,commonsense,adam3d},  predicting frame ordering  \cite{lee2017unsupervised}, 
predicting spatial context \cite{DBLP:journals/corr/DoerschGE15,pathakCVPR16context}, predicting color of  grayscale images \cite{DBLP:journals/corr/ZhangIE16}, predicting color of future video frames \cite{DBLP:journals/corr/abs-1806-09594},  predicting egomotion \cite{DBLP:journals/corr/JayaramanG15,DBLP:journals/corr/AgrawalCM15}, and many others. 
Our 2D-to-3D image encoder builds upon works that train 3D visual representations---instead of 2D---using view regression and contrastive view prediction as the pretext tasks \cite{commonsense,adam3d}. While \citet{adam3d} demonstrates the usefulness of such pretraining for 3D object detection, our work shows we can use the learned features to detect and associate objects, and infer 3D poses between objects. The work of \citet{pmlr-v87-florence18a} used  intra-scene correspondences provided by triangulation to train 2D CNNs for point feature matching. 
\citet{pot2018selfsupervisory} uses supervision from depth, egomotion and a vanilla 2D object detector to collect multiview images of the same object and learns  
2D feature representations  that cluster into discrete object identities.  
We consider a supervision setup similar to \citet{pot2018selfsupervisory}, but we pursue 3D feature representations.  We learn 3D object detection and pose estimation of objects, as opposed to solely 2D object detection.

\noindent \normalfont \textbf{Inverse graphics, analysis-by-synthesis.}
Approaches on inverse graphics or analysis-by-synthesis  attempt to map 2D images to complete 3D scene representations in terms of object 3D meshes, camera pose and scene layout \cite{conf/cvpr/KulkarniKTM15,Romaszko_2017_ICCV,DBLP:journals/corr/IzadiniaSS16}. 
Many works show they can recover parametrized 3D meshes or binary voxel occupancies of objects from videos and scenes \cite{DBLP:journals/corr/TulsianiZEM17,DBLP:journals/corr/NovotnyLV17a,3Dinterpreter,DBLP:journals/corr/TungHSF17} by unsupervised rendering and matching  to input depth maps.  Our work differs in that we pursue feature-based 3D representations instead. As such, we do not need to know a low parametric model of the object mesh ahead of time as in recent works \cite{mocap,conf/cvpr/KulkarniKTM15}, and we do not need a predefined set of 3D object shapes as in recent works \cite{Romaszko_2017_ICCV,DBLP:journals/corr/IzadiniaSS16}.

\noindent \normalfont \textbf{Few-Shot Object Recognition.} Existing work has proposed models that can learn to detect new objects with one or a few samples. 
However, these models cannot estimate 3D pose.
Metric-based few-shot learning approaches \cite{Koch2015SiameseNN, DBLP:journals/corr/VinyalsBLKW16, DBLP:journals/corr/SnellSZ17} learn an embedding space in which objects of the same category are clustered together in the latent space. After training, the model can infer the most similar instance from the support samples by comparing instances in the learned embedding space. However, there is no obvious method to directly use the embedding space to infer the relative object poses between the query instance and the support sample. 
Moreover, the learning of these models also relies on human labels: all approaches require standard few-shot learning dataset which consists of multiple sub-groups of images that are labelled as ``belonging to the same category''. The recently proposed method of \citet{Tian2020RethinkingFI} learns a classifier on top of supervised or self-supervised representations with few labels, and this outperforms previous few-shot learning approaches, but again it remains unclear how we can use the learned representations or the classifier to infer relative object poses.

\noindent \normalfont \textbf{Self-paced learning.} 
Many techniques in semi-supervised or unsupervised visual learning iterate between pseudo-label inference and classifier/feature  update using the inferred labels, in an Expectation-Maximization (EM) style algorithm \cite{soviany2019curriculum,DBLP:journals/corr/abs-1810-07911,DBLP:journals/corr/XieGF15,Shen_2019_CVPR}, yet existing work focus on improving 2D detection without considering detection in 3D.
For example, successful recent methods for domain adaptation \cite{soviany2019curriculum,DBLP:journals/corr/abs-1810-07911}  iterate between pseudo pixel label inference in the target domain and updating the pixel labellers.
These methods show the classifiers or detectors can improve without drifting. 
While our work self-infers cross-scene 3D correspondences to improve the features and infers pseudo 3D box labels to improve the 3D object detector, previous works operating in 2D image space do not consider this.

%% file: 4_model.tex
\section{3D Quantized-Networks (3DQ-Nets)}
\begin{figure*}[h!]
 \centering
 \includegraphics[width=1.0\textwidth]{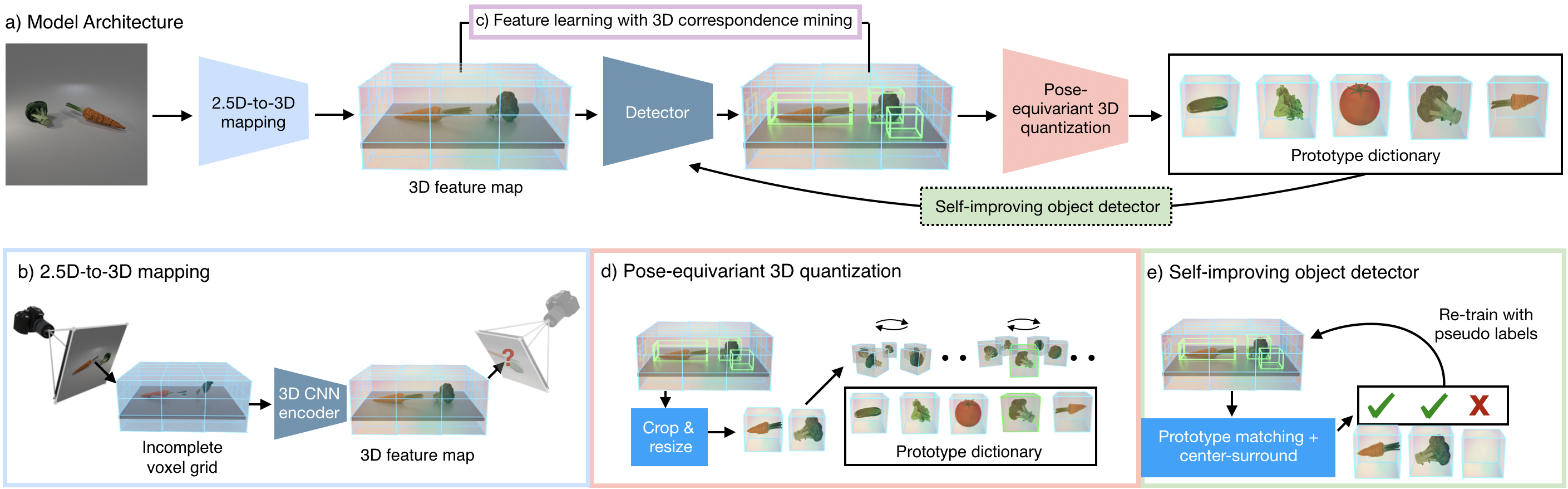}
 \caption{ Architecture for \textbf{3D Quantized-Networks (3DQ-Nets)}. 
 Given multi-view posed RGB-D images of scenes as input during training, our model learns to map a single RGB-D image to a completed scene 3D feature map at test time, by training for view prediction  (b). The model additionally uses cross-scene and cross-object 3D correspondence mining and metric learning, to make the features more discriminative (c).  Finally, using these learned features, our model quantizes object instances into a set of pose-canonical 3D prototypes using rotation-aware matching (d). These learned prototypes help improve our object detector by providing confident positive 3D object box labels (e) .
 }
 \label{fig:model}
\end{figure*}

We depict the architecture of our model in Fig.~\ref{fig:model}. 
Given a set of posed RGB-D images of a static scene, our model constructs a 3D scene feature representation by neurally lifting and registering features extracted from each frame using geometry-aware inverse graphics networks (GIGNs) (Sec.~\ref{sec:enc}). 
Our model detects objects in the inferred 3D scene representation (Sec.~\ref{sec:detect}) and matches the 3D object feature tensors against a set of 3D prototypes by searching over 3D rotations (Sec.~\ref{sec:quantization}). 
Concurrently, our model uses the detected 3D boxes to improve the 3D visual feature representation by iteratively inferring 3D part correspondences across objects detected in different scenes, and using metric learning to supervise the feature representation to reinforce the inferred correspondences (Sec.~\ref{sec:crossviewpred}). 

Our model iteratively optimizes over weights of the encoder, decoder, 3D detector module and prototypes, and uses individual modules to bootstrap the learning of the others. We pretrain the weights of the encoder and decoder of GIGNs by view prediction. We detail each module in their respective section and present the learning of the model in Sec.~\ref{sec:learning}.


\subsection{2.5D-to-3D lifting using Geometry-aware Inverse Graphics Networks (GIGNs)}
\label{sec:enc}
Geometry-aware Inverse Graphics Networks (GIGNs) \cite{commonsense,adam3d} ``lift" RGB-D images of static world scenes to 3D scene feature maps. The networks can be optimized end-to-end for a downstream task, such as supervised 3D object detection or unsupervised view prediction. To obtain the 3D scene feature maps, GIGNs are equipped with a differentiable 2D-to-3D inverse projection operation that can transform 2D feature maps into 3D feature maps.
We will denote the 3D feature map inferred from an input RGB-D image $I$ as 
$\M = \Enc(I) \in \mathbb{R}^{w \times h \times d \times c}$
where $w,h,d,c$ denote the width, height, depth and number of channels, respectively. Our experiments use $(w,h,d,c) = (72,72,72,32)$. 
GIGNs explicitly rotate and translate the feature maps inferred from different RGB-D views using their corresponding ground truth camera poses. As a result, feature maps from different views are all aligned to a common coordinate system. 

\paragraph{3D feature learning by predicting views}

We pre-train the encoder and decoder of GIGNs by predicting views using our posed RGB-D multiview image set. 

Following the work of \cite{commonsense, adam3d}, we train GIGNs to predict a query view given a single view input, which enforces the model to complete the missing or occluded information from the image.
Specifically, 
to predict a novel view, the scene feature map $\M$ is oriented to a sampled query viewpoint $v_{q}$ and decoded to an RGB image and occupancy grid, and then compared with the ground truth RGB ($I_{q}$) and occupancy ($O_{q}$) respectively:
\begin{align}
\begin{split}
\mathcal{L}^{v} =
&\|\Dec^{\text{\tiny RGB}}(\M,v_{q})-I_{q}\|_1 \\
&+ \log(1+\exp(-O_{q} \cdot \Dec^{\mathrm{occ}}(\M,v_{q}))),
\label{eq:viewpred}
\end{split}
\end{align}
\noindent 
The RGB output is trained with a regression loss, and the occupancy is trained with a logistic classification loss. Occupancy labels are computed through raycasting, similar to \citet{adam3d}. 
Please refer the supplementary material for more details. 

\paragraph{3D object detection}
\label{sec:detect}
A 3D detector operates on the output of the geometric encoder $\Enc$ and predicts a variable number of object boxes with associated confidences: $\mathcal{O}=\Det(\M) \in \{ (\hat{b}_{loc}^o, c^o) | \hat{b}_{loc}^o \in \mathbb{R}^6, c^o \in [0,1] \}$.
We follow the architecture used in the work of \citet{commonsense} for our detector.
We provide our detector with a ``warm start'' by pre-training it with 3D box annotations computed from triangulated 2D category-agnostic proposals from a publicly-available 2D objectness detector \cite{wu2019detectron2}. A detector trained with noisy annotations obtained from triangulation is expected to perform poorly, but it is sufficient for our system to start learning something useful. 
In Sec.~\ref{sec:learning} we describe our method for self-training the detector, so that it gradually learns to outperform its initialization.

\subsection{Quantizing objects into prototypes}
\label{sec:quantization}
Our model learns a set of 3-dimensional prototypes $\ee_k \in \mathbb{R}^{w_p \times h_p \times d_p \times c }, k \in \mathcal{K} = \{1, \ldots, K\}$ by clustering cropped 3D feature maps. Each prototype represents a set of similar objects.
The prototype serves as the cluster center of the set. To learn them, our model clusters objects in the scene in a pose-equivariant and scale-equivariant manner: similar object instances that vary in scale and pose are mapped to the same prototype. 
We crop the 3D scene feature map $\M$ given a detected box to obtain object 3D feature tensors, and resize it to match the common size of the 3D prototypes $\M^o=\resize(\crop(\M,b^o),[w_p,h_p, d_p])$. Our experiments use $(w_p,h_p,d_p) = (16,16,16)$.
We match detected objects' 3D feature tensors to prototypes using a rotation-aware feature matching. Specifically, we exhaustively search across rotations $\mathcal{R}$, in a parallel manner, considering increments of $10\degree$ along the vertical axis: 
\begin{align}
&(\zido,\zRo)= \argmin_{\substack{k \in \mathcal{K}, \RR \in \mathcal{R}}} \|\ee_k - \Rotate \left(\M^o,\RR \right)\|_2, \forall o \in \mathcal{O},
\label{eq:argmin}
\end{align}
where $ \Rotate \left(\M,\RR \right) $ explicitly rotates the content in feature map $\M$ with angle $\RR$ through trilinear interpolation.
Having assigned objects to oriented prototypes, we update our prototypes to minimize their Euclidean distance to the assigned oriented and scaled object tensors: 
\begin{align}
\mathcal{L}^{3DQ}(\ee)=&\sum \limits_{o=1}^{|\mathcal{O}|} \| \ee_{\zido} - \Rotate(\M^o, \zRo) \|_2
\label{eq:quant}
\end{align}
We initialize our prototype dictionary with a set of exemplars. To ensure prototype diversity at this initial stage, we build the dictionary incrementally, and only use an exemplar as a prototype if its feature distance to the already-initialized prototypes is higher than a threshold. 
Equations \ref{eq:argmin} and \ref{eq:quant} can be seen as expectation maximization steps iterating between exemplars-to-prototypes assignment and prototype updates.


\subsection{Cross-scene 3D correspondence mining}
\label{sec:crossviewpred}
Whether the model can establish the correct correspondence between objects and learn meaningful clusters relies on the quality of the visual features. To improve the visual features our model exploits visual similarity not only within scenes, but also across scenes.
While the view prediction objective of Eq.~\ref{eq:viewpred} exploits different views of the {\it same} scene to learn the features, 
our model further exploits part-based correspondence between objects in {\it different} scenes to further improve the learned features.
 We adopt the correspondence mining method of ArtMiner \cite{Shen_2019_CVPR} to operate in 3D as opposed to 2D:
Part based correspondences are hypothesized within detected objects and are verified by voting of their surrounding context voxels. If the original match is verified, hard-positive matches are then suggested in the surrounding of the match. Using the mined hard positive matches and randomly sampled negatives, we finetune the weights of our encoder $\Enc$ using metric learning. We empirically found that training with such cross-scene part-based correspondences helps improve the features.
We provide the implementation details in the supplementary material. 


\subsection{Iterative learning of object detection, visual features, and clustering}
\label{sec:learning}
Since the initialized object detector is sub-optimal due to the lack of groundtruth 3D boxes and can affect the rest of the modules, it is critical that we have a mechanism to improve it over time. To achieve this, we iterate our model over the following steps: 
(i) 3D object detection (Section \ref{sec:detect}). This generates a set of 3D object proposals.
(ii) Cross-scene object part correspondence mining and learning (Section \ref{sec:crossviewpred}). This updates the encoder weights $\Enc$ using metric learning on inferred cross-scene correspondence on the detected objects. 
(iii) Prototype update (Section \ref{sec:quantization}). This assigns detected object instances to prototypes and updates the prototypes $\ee$ by backpropagating the clustering loss in Eq. \ref{eq:quant}. 
(iv) Object detector update. We label 3D object proposals as positives or negatives using a combination of 3D center-surround saliency score and matching to prototypes score. After the object detector is updated, we can iterate from step one to improve the rest of the modules.

Specifically, we keep the 3D object proposals that have a good matching score against the learned prototypes and 
 discard the 3D object proposals whose 3D center-surround feature match score is below a threshold. The intuition is trust detection that either detects something that occurs often or has high saliency score.
 Center-surround saliency heuristic is used by numerous works for 2D and 3D object detection \cite{klein2011center,ju2015depth}. 
We then train the 3D object detector module to emulate such labels through standard gradient descent. 
In Fig.~\ref{fig:det_quali}-(a), we visualize the self annotations and improvement made by our self-improving detector over 4 iterations.

%% file: 5_exps_2.tex
\section{Experiments}

We test our framework in a variety of simulated environments and real world scenes. In simulation, RGB, depth and egomotion are provided by the simulator, whereas in the real world, RGB and depth are provided by Kinect sensors and egomotion is computed using camera calibration. 
Our experiments aim to answer the following questions:
\begin{enumerate}


 
    \item Do 3DQ-Nets recognize  objects better than CNN models pretrained on large labelled image datasets?
   \item How does the proposed pose-aware 3D clustering  compare against 2.5D pose-aware clustering,  3D pose-unaware clustering, or raw 3D point cloud registration?
    \item Does cross-scene 3D correspondence mining  improves  features over view-predictive training, and how much?
    \item In 3DQ-Nets, do feature learning, object clustering to prototypes, and 3D object detection improve  over training iterations?

\end{enumerate}

We benchmark our model on three datasets: (i) \textbf{CLEVR veggie} dataset: we build upon the CLEVR dataset \cite{johnson2017clevr} and add 17  vegetable object models bought from Turbosquid.
(ii) \textbf{CARLA} dataset: we created scenes using all 26  vehicle categories available in the CARLA simulator of Dosovitskiy \textit{et al.} \cite{Dosovitskiy17} 
(iii) \textbf{BigBIRD \cite{Singh2014BigBIRDAL}}: a publicly available dataset that contains multiview shots for 125 different objects rotating on a table. We assign the objects to 41 different object categories, combining similar objects into a single category. 

We further qualitatively evaluate our model on two  datasets:  
(iv) \textbf{Replica \cite{replica19arxiv}} dataset: we render images from the indoor meshes provided by Replica in AI Habitat simulator \cite{DBLP:journals/corr/abs-1904-01201}. The views are selected by moving the agent around randomly selected objects.
(v) \textbf{Real world desk scenes dataset}: training setup consists of 8 Kinect sensors surrounding the table to capture multiview RGB-D data. During test time, we only use a single Kinect sensor.
More details on our dataset collection are included in the supplementary material.

\subsection{Few-shot object category labelling}
\label{sec:fewshotlabellingsec}

In this experiment, we use ground-truth 3D bounding boxes during training of our model to isolate errors caused by the 3D object detection module. Out task is to classify object-centric image crops into object categories, when supplied with only two labelled object-image crop per category. This means, that e.g., in the CARLA dataset, we use 52 labelled object image crops. Note, the objects can be at any orientation. 
We evaluate the ability of our model and baselines to retrieve objects of the same category when supplied with these few labelled examples.

Given an annotated instance, our model finds the prototype that has minimum rotation-aware feature distance to the object instance, and  it propagates the label to all the  instances that are assigned to the same prototype. 
If a prototype is matched with more than one label, then the label which has matched the most is assigned to the prototype. 
Note that the small labelled set is not used to update our features or prototypes. 
In Table \ref{tab:fewshot}, we compare 3DQ-Nets against two 2D baseline models using pretrained ResNet-18 on ImageNet as their backbone: (i) Finetuning the top layer of
ResNet-18 with our training examples (ResNetClass), (ii) using the top average pool layer activations of ResNet-18 to retrieve and copy the label of the nearest neighbor instance from the training examples  (ResNetRet), i.e., not finetuning at all the weights.  
We show the results in Table \ref{tab:fewshot}. Our model outperforms both  ResNetClass and ResNetRet. Despite the fact the ResNet features are pre-trained on a large set of annotated images, our model can self-adapt in the new domain of each dataset, and thus learn more meaningful object distances, captured in the inferred 3D feature representations. 
On CLEVR-veggie dataset, ResNetRet performs slightly better than 3DQ-Nets. We suspect this is because the object categories in CLEVR-veggie appear in ImageNet, so the ImageNet pertaining likely provides discriminative features for these objects. 

\begin{table}[t!]
\begin{footnotesize}
\centering
\setlength{\tabcolsep}{0.5em} 
\renewcommand{\arraystretch}{1.1}
\begin{tabular}{|c|c|c|c|c|}
\hline
{Datasets.} & ResNetRet & ResNetClass & 3DQ-Nets \\ \hline
{CARLA} &  0.27 & 0.58 & \textbf{0.71} \\ \hline
{CLEVR} & \textbf{0.80} & 0.72 & 0.75 \\ \hline
{BigBIRD} &  0.40  & 0.67 & \textbf{0.82}  \\ \hline
\end{tabular}
\caption{
\textbf{Few shot object category labelling accuracy}
}

\label{tab:fewshot}
\end{footnotesize}
\end{table}

\subsection{Clustering with 3D pose-aware quantization}
\label{sec:cluster3d}

\begin{figure}[h!]
    \centering
    \includegraphics[width=0.45\textwidth]{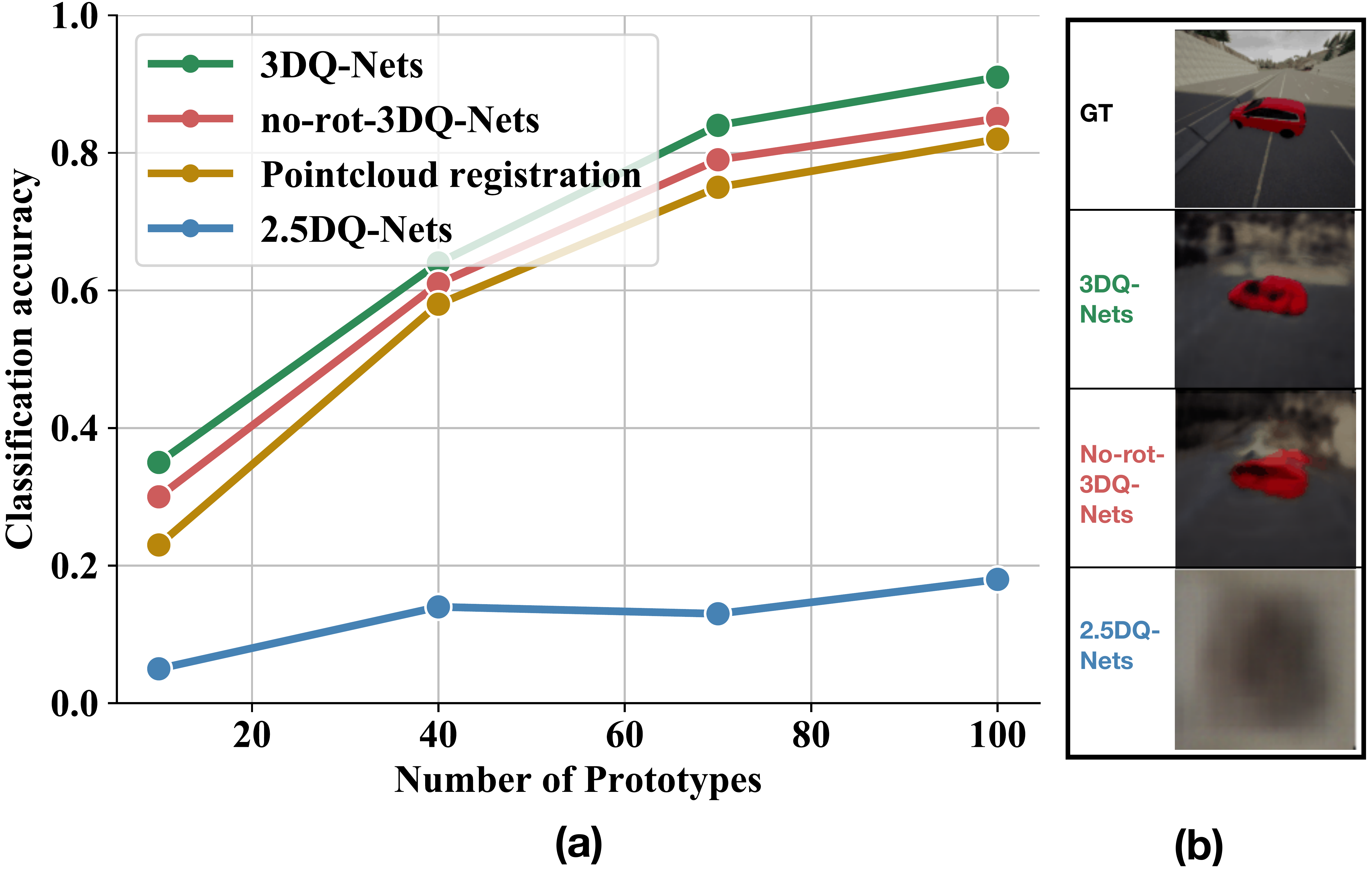}
      \caption{
      \textbf{(a) Unsupervised classification accuracy} with varying length of prototype dictionary in CARLA.
\textbf{(b) Scene reconstruction} results using the learned prototypes from our model and the baselines.
    }
    \label{fig:acc_neuralrender}
\end{figure}

In this experiment, we evaluate the importance of  3D pose-aware quantization for 3D object clustering. We compare our model against three baselines: (i) 2.5DQ-Nets, a 2D CNN model that takes concatenated RGB and depth as input and  quantizes detected 2D image patches into a discrete set of 2D prototypes by optimizing an autoencoding objective. During quantization, the model conducts 2D rotation search. (ii) no-rot-3DQ-Nets, a model similar to ours except that it assigns instances to 3D prototypes without rotation search. 
(iii) Pointcloud registration \cite{mitra2004registration}, a method that uses registered point clouds as prototypes and conducts 3D rotation aware search to infer the identity of the closest 3D poincloud prototype and the 3D pose of the object instance with respect to the prototype. 
For our model and baselines we consider ground-truth 3D  and 2D object boxes to isolate the error from different detectors. 


To evaluate the unsupervised classification accuracy using prototypes, we use LIN-MATCH, a bipartite graph matching method \cite{kuhn1955hungarian}, that finds the permutation of prototype indices that minimizes the classification error.
We show these comparisons with varying length of the prototype dictionary
in Figure \ref{fig:acc_neuralrender} (a). 
We see in Figure \ref{fig:acc_neuralrender} (a) that models that use 3D representation achieve significantly higher accuracy compared to models using 2D representation. Further adding rotation search in 3D during clustering improves the performance since the operation enforces objects with similar appearance but with different poses to be clustered together. We also show that being able to inpaint objects from a single view during inference helps our model in outperforming the Pointcloud registration baseline that needs to handle incomplete input object-centric poinclouds. In Figure \ref{fig:acc_neuralrender} (b) we show the scene reconstruction results of our models and the neural baselines after replacing the object in the scene with its best matched prototype under the inferred pose and rendering the 3D feature map through the learned decoder.  Please refer the supplementary material for more details on the baselines and results on other datasets.

\subsection{3D feature learning with 3D correspondence mining}
\label{sec:replearn}

In this experiment, we evaluate  the contribution of  3D mining  in  feature learning, by evaluating the features in object category few shot retrieval. We compare it against the following feature learning methods: (i) Resnet-18 pretrained on Imagenet dataset (ResNet), where we average-pool features within the projected (ground-truth) 2D object boxes to represent the objects. (ii) GIGNs trained with RGB view and occupancy prediction (rgb-occ) of \cite{commonsense}. (iii) GIGNs trained with object-centric view contrastive prediction (rgbocc+VC) of \cite{adam3d}. (iv) We improve (iii) by using the same metric learning loss function \cite{he2019momentum} as our model (rgbocc+VC*). (v) GIGNs trained additionally with cross-scene 3D mining (ours). For (ii),(iii),(iv),(v), we use the cropped 3D feature maps from 3D object boxes to represent the objects. 
We randomly sample 1000 objects and retrieve their nearest neighbors by considering the maximum inner product across 36 rotations against a pool of another 1000 objects. For (i), we consider 2D rotation search as opposed to 3D. 
We show category-level retrieval precision within the first 10 retrieved nearest neighbors (i.e., precision@10) in  Table \ref{tab:retrieval}. 

  As shown in Table \ref{tab:retrieval}, cross-scene correspondence mining improves the retrieval results. In the CLEVR dataset, ResNet outperforms our model. 
  Our model performs the best among the unsupervised methods. 

\begin{table}[t]
\begin{footnotesize}
\centering
\setlength{\tabcolsep}{0.5em} 
\renewcommand{\arraystretch}{1.1}
\begin{tabular}{|c|c|c|c|c|c|c|}
\hline
{Datasets} & ResNet   & rgbocc & rgbocc+VC  & rgbocc+VC*  & ours \\ \hline
{CARLA} & 0.49 & 0.62 & 0.55  & 0.67  & \textbf{0.80} \\ \hline
{CLEVR} & {\bf 0.87} & 0.74 & 0.71  & 0.74 & 0.81 \\ \hline
{BigBIRD} & 0.47 & 0.44  & 0.69  & \textbf{0.77} & 0.73\\ \hline
\end{tabular}
\caption{
\textbf{Retrieval results (precision@10 nearest neighbors)} for different architectures and objectives for 2D and 3D visual representation learning.} 
\label{tab:retrieval}
\end{footnotesize}
\end{table}

\subsection{Joint training of 3D object detection, feature learning and clustering}

\begin{table}
\begin{footnotesize}
\centering
\begin{tabular}{|l|*{4}{c|}}\hline
\backslashbox{Task}{Iterations}
& Iter 0 & Iter 1 & Iter 2 & Iter 3\\\hline
Feature Learning &0.72 & 0.76 & 0.79 & 0.79\\\hline
Quantization & 0.51 & 0.63 & 0.65 & 0.66\\\hline
Detection & 0.43 & 0.48 & 0.51 & 0.52\\\hline
\end{tabular}
\caption{Performance across training EM iterations of our model in CLEVR. Feature learning is measured using the same technique as Table \ref{tab:retrieval}. Object quantization uses the same measurement technique as Fig. \ref{fig:acc_neuralrender} (a). Detection performance is measured by meanAP at IoU $= 0.5$.}
\label{table:continual_learn_main}
\end{footnotesize}
\end{table}

\begin{figure}[h!]

\begin{center}
\includegraphics[width=0.47\textwidth]{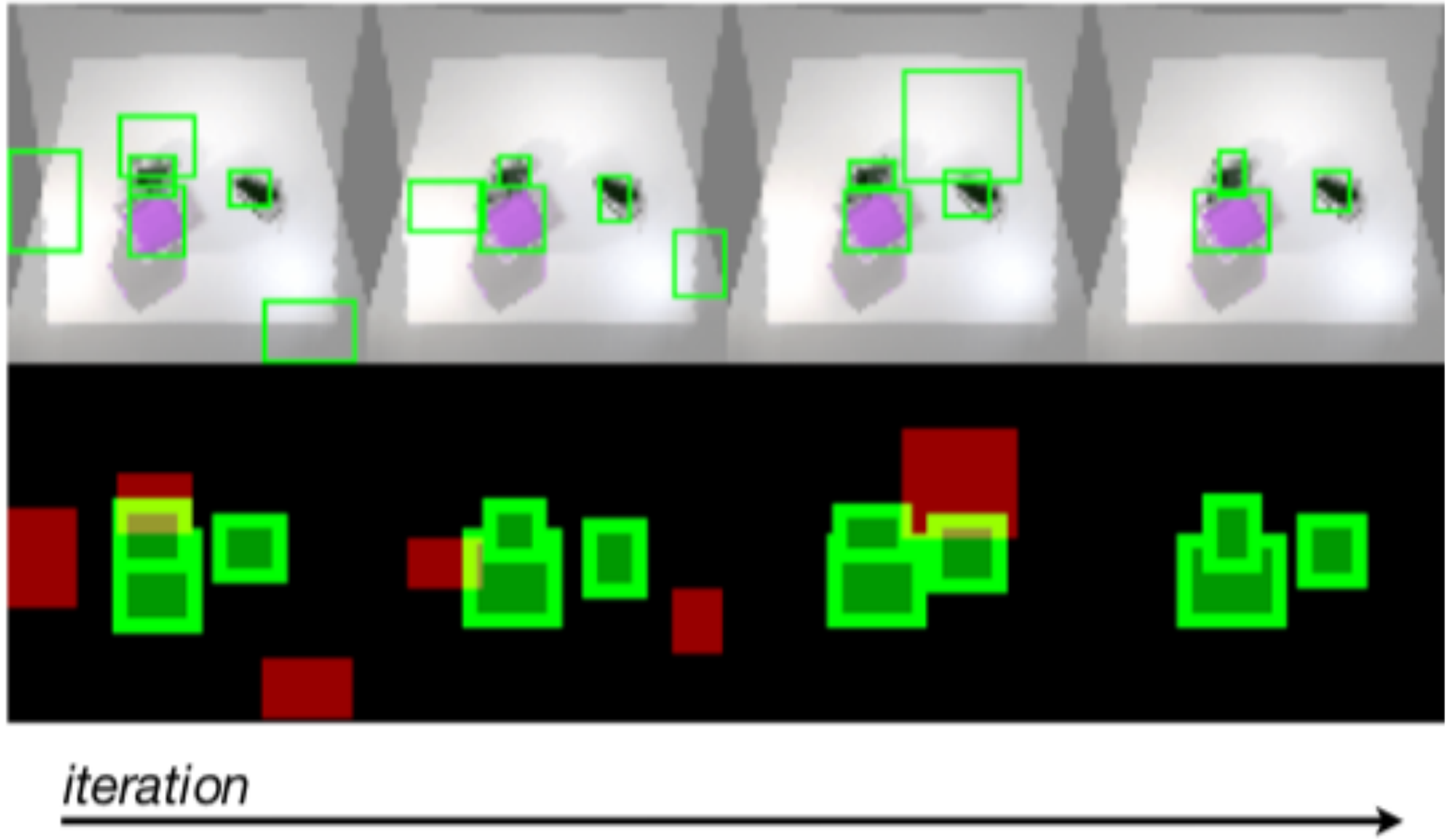}
\end{center}
\caption{ \textbf{Detection improvement over 4 iterations.} The first row shows the input image and the proposals of the object detector. The second row shows the annotations assigned to the proposals using the 3D prototype distance and 3D center-surround score. We show that our detector improves over time without any ground truth 3D proposals. }
 \label{fig:det_quali}
\end{figure}

Table \ref{table:continual_learn_main} shows evaluations of our different modules during  4 iterations of EM. We see that the performance of all our modules improves over iterations. To initialize the weights for the modules (Iteration 0), we warm-start the 3D scene features using RGB view and occupancy prediction (rgbocc), and use the 3D object proposals provided by triangulated 2D boxes from 2D objectness detector to  train the detector, visual features and prototypes.
From Iteration 1 onwards, we use the 3D detected boxes from the trained detector as inputs, and use 3D mining to update the features. We subsequently improve the detector and the rest of the modules iteratively.
We show that all modules can boostrap one another and continually improve over iterations. We further show our detector improvement over time in Figure \ref{fig:det_quali}.

\subsection{Scene parsing using prototypes}

\begin{figure}[h!]
    \centering
    \includegraphics[width=0.45\textwidth]{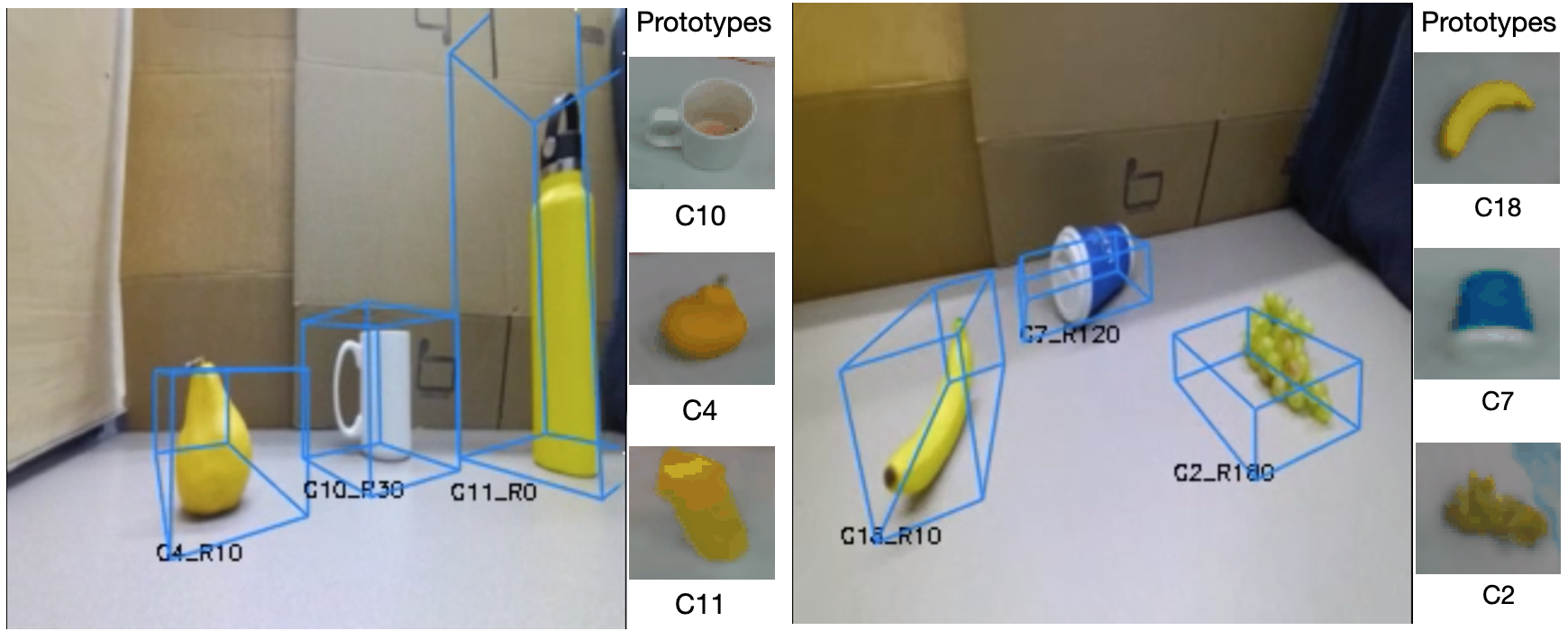}
  \caption{\textbf{Real world scene parsing results}. More results are available on the project webpage.}
  \label{fig:realworldfig}
\end{figure}

Our learnt prototypes capture each object instance in its canonical pose. We use these prototypes for task of scene parsing. Given a new scene, we first detect all the objects and extract their features from the scene.
Then, we match the object-centric feature maps with all the prototypes using a rotation aware similarity check explained in Section \ref{sec:quantization}. For each detected object instance in the scene, we visualize the matched prototype number (C) and the respective rotation angle along vertical axis (R) as seen in Figure \ref{fig:realworldfig}. We also visualize the respective prototypes by neurally rendering them to images. Please refer supplementary material for the video on real world scene parsing. Results on Replica dataset are provided in the supplementary material.

%% file: 6_conclusion.tex
\subsection{Limitations}

The presented framework currently has the following limitations:
\textbf{(i)} 
Prototypes do not deform. A prototype is rotated and scaled, but not non-rigidly deformed. Adding such deformable parametrization \cite{DBLP:journals/corr/abs-1708-04672} would increase the expressiveness of the prototype dictionary.
\textbf{(ii)} 
Prototypes cannot be stylized. Allowing prototypes to be stylized by changing their appearance using predicted stylization parameters \cite{DBLP:journals/corr/HuangB17}, would again increase their expressiveness dramatically. 
\textbf{(iii)} 
The model cannot learn from videos of dynamic scenes, i.e., scenes with independently moving or deforming objects. Overcoming this limitation would require tracking the moving objects over time.

\section{Conclusion}
We presented a system that given multi-view posed 2.5D images learns to detect the objects in 3D and organizes them into a set of 3D prototypes in their canonical poses and scales. 
We applied our method to various datasets in simulation and in the real world. We demonstrated the usefulness of our framework in few shot learning,
where prototypes propagate a small number of semantic labels to object instances. In that task, our model outperforms ImageNet-pretrained and 2.5D feature learning baselines. We further empirically showed that the modules of our framework improve over time, the proposed 3D prototypes are more expressive than 2.5D and registered point cloud equivalents, and  3D cross-scene correspondence mining dramatically improves retrieval accuracy, compared to view prediction objectives alone. 
Learning from videos of dynamic scenes and incorporating deformability and stylization during prototype matching and learning, as discussed in the limitations section, are clear avenues for future work. 


%% file: 9_suppl.tex
\section{Appendix Overview}
\label{sec: append_overview}
In Section \ref{sec:supp_data_prep}, we provide details for all our datasets and their collection process. In Section \ref{sec:imp_details}, we provide implementation details for each of our modules and we also provide further implementation details of the baselines.
In Section \ref{sec:added_results}, we provide additional qualitative/quantitative results on Replica and other datasets.


\section{Dataset preparation}
\label{sec:supp_data_prep}

\paragraph{\normalfont \textbf{CLEVR Dataset.}} 
We build upon the CLEVR Blender simulator \cite{johnson2017clevr} and add 17 vegetable object models bought from Turbosquid \cite{turbosquid}, in addition to the object models available in CLEVR. So in total our dataset has 41 unique object models. We consider each object model to be a separate object category, this information will be used for evaluation purposes, not at training time. We create scenes as follows:  Each object model is randomly rotated ($0\degree$ to $360\degree$ along vertical axis), translated (randomly within a sphere of radius 10.5 units) and scaled (0.75 to 1.25 times the actual size). Each scene contains up to 3 objects. We  randomly vary the lightning of each scene.We render each scene by placing 28 RGB-D cameras at elevations ranging from $26\degree$  to $80\degree$ with $13\degree$ increments and azimuths ranging from $0\degree$ to $360\degree$  with $45\degree$ increments. Each camera is placed within a sphere of radius 1.5 metres from the center of the scene.

\paragraph{\normalfont \textbf{CARLA Dataset.}} 
Our CARLA dataset uses the 26  vehicle classes available in the CARLA simulator. We consider each vehicle model to be a separate object category, again this information will be used for evaluation purposes, not at training time. 
Each rendered scene consists of either one or two vehicles. Each scene in the training set consists of multi-view RGB and depth images of static vehicles placed at randomly selected spawn points. We generate scenes by randomly selecting a map from the available CARLA maps. Then we perturb the weather conditions randomly by setting cloudiness to a value in [0, 70], precipitation to be within [0, 75], and sun\_altitude\_angle to be within [30, 90]. For single-vehicle scenes, we randomly select a spawn point and place a vehicle at that spawn point. Then we place 17 RGB-D randomly cameras around the vehicle.  The origin of the vehicle serves as the origin with respect to which the extrinsic matrices of all the cameras are calculated. For vehicles in CARLA, the x axis points forward, y axis points to the right and the z axis points upwards. We place the first eight RGB-D cameras on the boundary of a circle centered at the vehicle's origin with radius=3.4m ,height z=1.0m and with yaw angle varying from $-40\degree$ to $-285\degree$ with increments of $-35\degree$ each. The next eight RGB-D cameras again follow the same setup but with z=3.0m. Finally, the last RGB-D camera is placed overhead with z=5.0m and pitch=$-90\degree$.
\\
For two-vehicle scenes, we first place the first vehicle at a randomly selected spawn point. We then select another spawn point from nearby spawn points and position the second vehicle there. This is required so that we can have both vehicles in the field of view of majority of the cameras. The origin in two vehicle setup is taken to be the mean of the origins of the two vehicles. All camera extrinsic matrices are calculated with respect to this origin. We again randomly place 17 RGB-D cameras around the origin. The first camera is placed at x=4.5m, z=1m, and yaw=-180. The next seven cameras are placed on the boundary of a circle of radius 7.5m, height 5.5m, and centered at the origin. The yaw angle is varied from $-40\degree$ to $-285\degree$ (with the exception of $-180\degree$) with increments of $-35\degree$.
The next seven cameras are placed on the boundary of a circle of radius 4.5m, height 6.5m, and centered at the origin. Each camera has pitch $-40\degree$. The yaw angle is varied from $-40\degree$ to $-285\degree$ with increments of $-35\degree$. The final camera is placed overhead with z=5m and pitch=$-90\degree$.

\paragraph{\normalfont \textbf{BigBIRD Dataset.}} 
BigBIRD dataset consists of 125 objects placed on a rotating table. We use the 'Raw-RGB-D' dataset provided by BigBIRD. The camera setup consists of 5 RGB-D cameras placed in an arc, with the first camera in front of the object and the last camera overhead. The cameras capture the RGB and depth images of the rotating object at every $3\degree$ interval. This gives us 600 RGB-D images for each object. For our use case, we treat the object as stationary and instead assume that there are cameras placed at every $3\degree$ interval capturing multi-view images of the static object. This setup results in 600 RGB-D cameras placed around a stationary object. We assign the 125 object classes to 41 different object classes, combining similar objects, e.g. clif\_crunch\_chocolate\_chip and clif\_crunch\_peanut\_butter, into a single class, thus satisfying our use-case. Note that category labels are used for evaluation purposes, not at training time. 

\paragraph{\normalfont \textbf{Replica Dataset.}} 
We render our Replica dataset by loading each of the 18 3D indoor meshes provided by Replica \cite{replica19arxiv} in AI Habitat \cite{DBLP:journals/corr/abs-1904-01201}. To generate scenes with objects, we pick some objects appearing in the vicinity to each other, move the agent around those objects and click 6 multi-view RGB-D images. For retrieval, compression, and detection tasks, we only consider objects which satisfy the following conditions: (1) They should be visible in at least 4 out of the 6 views. (2) The 2D bounding box for the object should have an area greater than 1000 $pixel^2$. (3) The ratio of the number of points occupied by an object in the semantic map to the area of the 2D bounding box for that object should be greater than 0.1. Our dataset collected from replica consists of 26 unqiue object categories.

\paragraph{\normalfont \textbf{Real world desk scenes dataset.}}
This dataset consists of a set of 18 different objects placed on a table seen by a  dome of 8 Microsoft Azure Kinect sensors. We know the intrinsics of the cameras and calculate the extrinsics by calibrating them using OpenCV's \cite{opencv} checkerboard caliberation technique. Since we cannot have annotations in the real world, we only show qualitative results for the model trained on this dataset. During test time, we collect RGB-D images of multiple objects placed on a table by moving a single Microsoft Azure Kinect around the scene. To get the camera extrinsics at different points of time, we estimate the trajectory of the camera by calculating its rigid body transformations using consecutive RGB-D images. The transformation is estimated using the point cloud matching technique described in \cite{park2017colored}. We utilise Open3D's \cite{Zhou2018} implementation of \cite{park2017colored}. Since the resolution of the RGB images and the depth maps are different, we create a RGB-D image by mapping the depth map to the RGB image to obtain the depth values for all the RGB pixels. This mapping is done using the linear interpolation module provided as part of the Kinect SDK.

\section{Implementation Details}
\label{sec:imp_details}

\begin{figure*}[h!]
    \centering
    \includegraphics[width=1.0\textwidth]{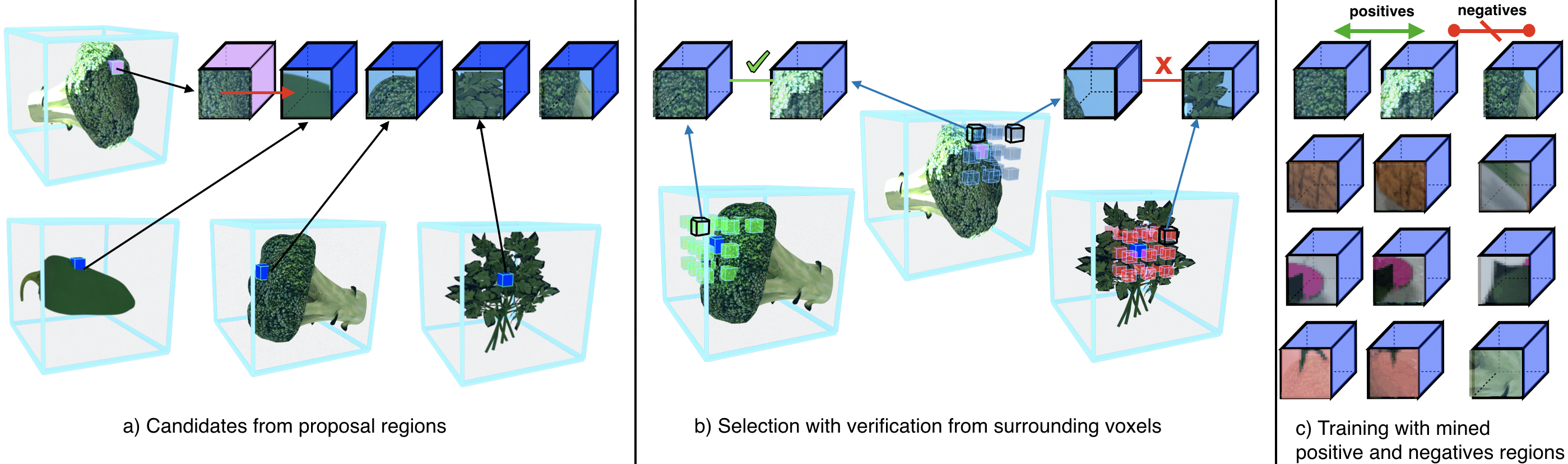}
      \caption{
      \textbf{Cross-scene 3D correspondence mining}. (a) We show that our approach relies on part-level correspondences obtained by matching the features of the query region (in pink) to a pool of  object-centric 3D features maps. (b) These part-level correspondences are verified based on how well their surrounding voxels match with one another in a spatially consistent manner. (c) Finally we train our 2.5D-to-3D lifting module by doing metric learning using the verified positive regions and randomly sampled negatives. 
    }
    \label{fig:3dmine}
\end{figure*}

\paragraph{\normalfont \textbf{Code, training details and computation complexity.}} Our model is implemented in Python/Pytorch. We keep our batch size as 2, our learning rate is kept as $10^{-3}$ for view prediction training and is dropped it down to $10^{-4}$ when training for all further tasks.  We use the Adam optimizer with $\beta_1 = 0.9$, $\beta_2 = 0.999$.
Our model takes 24hrs (approx. 100k iterations) of training for convergence and requires 0.8 seconds for an inference step on a single V100 GPU.

\paragraph{\normalfont \textbf{Inputs.}}
For all datasets, we resize input RGB images to a resolution of 256x256 pixels. We randomly select images from 2 views (query view and target view) of the multi-view scene as inputs for training our model, while we use a single view for testing.
\paragraph{\normalfont \textbf{Geometry-aware Inverse Graphics Networks(GIGNs)}}
Our 2.5D-to-3D lifting, 3D occupancy estimation and 2D RGB estimation modules follow the exact same architecture as \cite{adam3d}. We explain the implementation details of each of these modules below.
\\\\
\normalfont \textbf{2.5D-to-3D lifting}
Our 2.5D-to-3D unprojection module takes as input RGB-D images and converts it into a 4D tensor $\textbf{U} \in \mathbb{R}^{w \times h \times d \times 4}$, where $w,h,d$ is $72,72,72$. We use perspective (un)projection to fill the 3D grid with samples from 2D image. Specifically, using pinhole camera model \citep{hartley2003multiple}, we find the floating-point 2D pixel location that every cell in the 3D grid, indexed by the coordinate $(i,j,k)$, projects onto from the current camera viewpoint. This is given by $[u,v]^T = \K \S [i,j,k]^T$, where $\S$, the similarity transform, converts memory coordinates to camera coordinates and $\K$, the camera intrinsics, convert camera coordinates to pixel coordinates.
Bilinear interpolation is applied on pixel values to fill the grid cells. We obtain a binary occupancy grid $\textbf{O} \in \mathbb{R}^{w \times h \times d \times 1}$ from the depth image $\textbf{D}$ in a similar way. This occupancy is then concatenated with the unprojected RGB to get a tensor \begin{math}[\textbf{U}, \textbf{O}] \in \mathbb{R}^{w \times h \times d \times 4}\end{math}. This tensor is then passed through a 3D encoder-decoder network, the architecture of which is as follows: 4-2-64, 4-2-128, 4-2-256, 4-0.5-128, 4-0.5-64, 1-1-$F$. Here, we use the notation $k$-$s$-$c$ for kernel-stride-channels, and $F$ is the feature dimension, which we set to $F=32$.
We concatenate the output of transposed convolutions in decoder with same resolution feature map output from the encoder. The concatenated tensor is then passed to the next layer in the decoder. We use leaky ReLU activation and batch normalization after every convolution layer, except for the last one in each network. We obtain our 3D feature map $\textbf{M}$ as the output of this process.

\normalfont \textbf{3D occupancy estimation.}
In this step, we want to estimate whether a voxel in the 3D grid is ``occupied`` or ``free``. The input depth image gives us partial labels for this. We voxelize the pointcloud to get sparse ``occupied`` labels. All voxel cells that are intersected by the ray from the source-camera to each occupied voxel are marked as ``free``. We give $\M$ as input to the occupancy module. It produces a new tensor $\textbf{C}$, where each voxel stores the probability of being occupied. We use a 3D convolution layer with a $1 \times 1 \times 1$ filter followed by a sigmoid non-linearity to achieve this. We train this network with the logistic loss,
$\loss_\text{occ} = (1/{\sum \textbf{\^I}}) \sum \textbf{\^I} \log(1 + \exp(- \textbf{\^C} \cdot \textbf{C}) ),$
where $\textbf{\^C}$ is the label map, and $\textbf{\^I}$ is an indicator tensor, indicating which labels are valid. Since there are far more ``free'' voxels than ``occupied'', we balance this loss across classes within each minibatch.
\\
\normalfont \textbf{2D RGB estimation.}
Given a camera viewpoint $v_{q}$, this module projects the 3D feature map $\M$ to ``render`` 2D feature maps. To achieve this, we first obtain a view-aligned version, $\M_{v_{q}}$,  by resampling $\M$. 
The view oriented tensor, $\M_{v_{q}}$, is then warped so that perspective viewing rays become axis-aligned. This gives us the perspective-transformed tensor $\M_{proj_{q}}$.
This tensor is then passed through a CNN to get a 2D feature map $v_{q}$.
The CNN has the following architecture (using the notation $k$-$s$-$c$ for kernel-stride-channels): max-pool along the depth axis with $1 \times 8 \times 1$ kernel and $1 \times 8 \times 1$ stride, to coarsely aggregate along each camera ray, 3D convolution with 3-1-32, reshape to place rays together with the channel axis, 2D convolution with 3-1-32, and finally 2D convolution with 1-1-$E$, where $E$ is the channel dimension, $E=3$.

\paragraph{\normalfont \textbf{3D Detector.}}
Our detector follows the architecture design of  \cite{commonsense}, which extends the 2D faster RCNN  architecture to predict 3D bounding boxes from 3D features maps, as opposed to 2D boxes from 2D feature maps.  The detector takes the 3D feature map from the 2D-to-3D lifting as input to predict object bounding boxes. 
The detector consists of one down-sampling layer and three 3D residual blocks, each having 32 channels. We use 1 anchor box at each grid location in the 3D feature map with a size of 0.12 meters for the CLEVR dataset, and a size of 1.7 meters for the CARLA dataset. The detector will output an objectness score for each anchor box and select boxes that exceeds a threshold. We set the threshold to be 0.9. While training the 3D detector, we freeze the weights of the 2.5D-to-3D lifting module and only finetune the layers after the 3D feature maps. We empirically found that training both the detector and the 2.5D-to-3D features degrades the quality of the features and the detector. 
\paragraph{\normalfont \textbf{3D correspondence mining.}}
We randomly select 2000 object instances from our training data to create two
pools (Query Pool \& Target Pool) of size 1000 each. Each pool maintains object-centric 3D features of spatial size $16 \times 16 \times 16$
extracted from the 3D feature map using the detected boxes. As shown in Figure \ref{fig:3dmine} for each training iteration, we randomly select
a $2 \times 2 \times 2$ patch on an object sampled from the Query Pool, and by doing exhaustive search (across 36 different orientations along the vertical axis) and verification in the target pool object features we mine positive patches for metric learning training.

However, searching over all the possible patches (we extract $4 \time 4 \time 4$ patches from each object) for all 1000 objects in the target pool with all the 36 poses is computationally inefficient. To reduce computation, we first complete a rough search at the object-level to retrieve objects which are similar to the query object, then we do fine-grained search at the part-level by searching over possible patches from these objects of interest. We do this by ranking objects based on their cosine distance (we take the maximum cosine distance across 36 rotations) with the query object, and take only the top 30 objects 
to perform fine-grained search on the patch-level. 

For each target object, we extract $4 \time 4 \time 4$ patches to compare with the query patch. For each patch, we conduct a 
    spatial consistency check similar to the work of \cite{Shen_2019_CVPR}: instead of computing inner product between the patches, we compare the surrounding patches of these patches. We take the patches 6 unit Manhattan distance away from the patch center and compute an inner product on these surrounding patches. The summation of the inner product between all the surrounding patches serve as the final matching score for center patches. We take the top 200 patch retrievals based on the score, and take the 8 corners from their surroundings as positives. We create negatives by randomly selecting a pair of patches from the pool. However, training with naively sampled negatives on the fly is unstable. Following the suggestion from the work of \cite{he2019momentum}, we 
maintain a dictionary of size 100,000 for the negatives examples, and
do momentum update on our 2.5D-to-3D lifting module.

\paragraph{\normalfont \textbf{Quantizing objects into prototypes.}}
Each object prototype is a 3D feature tensor of size $16 \times 16 \times 16 \times 32$. We initialize these prototypes incrementally and assign an exemplar as a prototype only if its feature distance to the already-initialized prototypes is lower than a cosine distance of 0.8. This ensures diversity of prototypes during initialization. While associating exemplars to a prototype, we check over 36 different rotations along the vertical axis at $10\degree$ increments. We keep our prototype dictionary size $K$ as 50 for all the datasets. Empirically from Figure 3(a)(main paper) we have found that, K should be large enough to cover the object variability in the dataset.

\paragraph{\normalfont \textbf{3D object detection supervised by prototypes distance and center-surround score.}}

We initialize our 3D object detector by training it using triangulated 2D class-agnostic bounding box detections obtained from a 2D objectness detector. For our 2D objectness detector we use the publicly available code of \cite{wu2019detectron2} which uses a Faster R-CNN\cite{ren2015faster} backbone architecture and is trained using lots of 2D bounding box annotations from the COCO dataset \cite{cocodataset}. A 3D detector trained with noisy annotations obtained from triangulation is expected to perform poorly, and thus we found it critical that we have a mechanism to improve it over time. 

In order to improve our detector, we first crop the object features from the inferred 3D feature map using the predicted 3D bounding boxes of our detector and resize their spatial dimension to $16 \times 16 \times 16$ to obtain object-centric feature tensors. For every cropped object tensor we calculate the cosine distance which is maximum amongst all the  prototypes in the dictionary. If this calculated distance for a proposal is greater than 0.8 then we keep it as a valid proposal. In-order to find the invalid proposals we use 3D center-surround saliency. Specifically we calculate the  average cosine-distance of the cropped object tensor with its surrounding (top, down, left, right, front, behind) across all 3 axes. If the average cosine-distance is above 0.65 then we consider that proposal as invalid. We finally use the valid proposals as pseudo ground truths to further train the detector. We pass our gradients only through the aggregated region of all the valids and invalids, with the fear that there could be a prospective object proposal in the remaining region which was never predicted by the detector. The hope is that via iterative learning our detector learns 3D objectness and is thus able to get rid of it's bad proposals.

\paragraph{\normalfont \textbf{Implementation details for all baselines used in experiment subsection 4.2(main paper).}}
Inorder to ablate the learnt latent representation we make sure that the prototypes for our model and all our baselines use the same number of bytes.
\\
\textbf{Pointcloud Registration\cite{mitra2004registration}}
We specifically select this as one of our baselines inorder to compare our model against a traditional computer vision method which doesn't use deep learning to learn its features. In this baseline we use registered point clouds as prototypes. We conduct 3D rotation aware search to identify the identity and 3D pose of the new object instances with respect to the prototype. We voxelize the point clouds into a 3D grid while computing the cosine similarity between the two. Since the new object instance will have an incomplete point cloud, we compute similarity only for the occupied points of the new object instance.
\\
\textbf{2.5DQ-Nets}
This baseline has the exact architecture as our 3DQ-Nets model, except it uses 2D CNNs instead of 3D and uses 2D rotation aware search instead of a 3D search. The model is trained on autoencoding the same RGB-D view instead of different query view like our model.
\\
\textbf{no-rot-3DQ-Nets}
This baseline is an ablation for 3DQ-Nets rotation aware check. In this model we follow the same procedure as 3DQ-Nets but do not do any rotation aware check while matching 3D instances with the prototypes.
\section{Additional results}
\label{sec:added_results}
\subsection{Quantiative results for clustering with 3D pose-aware quantization}
Due to insufficient space in the main paper, in this section we further extend the experiments conducted in Figure 3(a)(main paper) on CARLA datset to all other datasets. In Table \ref{tab:unpclass} we show the unsupervised classification accuracy using the same testing/training setup of Section 4.2(main paper). For this experiment we set the number of prototyes (K) as 50.

\begin{table}[t!]
\begin{footnotesize}
\centering
\setlength{\tabcolsep}{0.2em} 
\renewcommand{\arraystretch}{1.0}
\begin{tabular}{|p{1.5cm}|p{1.5cm}|p{1.5cm}|p{1.5cm}|p{1.5cm}|}
\hline
{ Datasets.} &  2.5DQ-Nets  &  no-rot-3DQ-Nets &  Pointcloud registration & 3DQ-Nets \\ \hline
{CLEVR} &  0.23 & 0.73 & 0.51 & \textbf{0.77} \\ \hline
{BigBIRD} &  0.28  & 0.81 & 0.57 & \textbf{0.83} \\ \hline
\end{tabular}
\caption{\textbf{Unsupervised classification accuracy}  with dictionary size of 50 prototypes on CLEVR and BigBIRD datasets. }
\label{tab:unpclass}
\end{footnotesize}
\end{table}

\subsection{Quantiative results for 3D object detection improvement}

\label{sec:3d_obj_det_sup_by_comp}
In this section, we show how the mean average precision (meanAP) of our 3D detector  improves  over time when supervised by visual compression and 3D center-surround saliency. We consider two initialization schemes for our 3D detector: i) we train our detector with a set of ground-truth 3D bounding boxes in a training set (3D-pretrain), ii) we train our detector by triangulating 2D object proposals from our 2D objectness detector, as described in Section 3.4(main paper), again in a training set (2Dtriang-pretrain). We show results in Table \ref{tab:detimprove}. From the results, we see our detector can improve its detection by a large margin after finetuning its weights by learning on the positive examples suggested by the learned object prototypes and negatives examples from the center surround check. 
 \begin{table*}[t!]
 \begin{footnotesize}
\centering
\renewcommand{\arraystretch}{1.1}
\begin{tabular}{|c|c|c|c|c|}
\hline
{Datasets } & 3D pretrain &  3DQ-Nets (\textit{final}) & 2D triang-pretrain & 3DQ-Nets (\textit{final}) \\ \hline
{CARLA} &  0.41  & \textbf{0.59} & 0.32 & \textbf{0.41}\\ \hline
{CLEVR} &  0.42 & \textbf{0.61} & 0.37 & \textbf{0.52} \\ \hline
\end{tabular}
\caption{\textbf{Initial and final 3D detection meanAP} at IoU=0.5 using detected 2D proposal triangulation versus ground-truth 3D bounding boxes in a training set. In both cases, 3DQ-Nets improve the detector over time, supervised by compression and 3D center-surround saliency.}
\label{tab:detimprove}
\end{footnotesize}
 \end{table*}
 
\begin{figure*}[h!]
    \centering
    \includegraphics[width=0.95\textwidth]{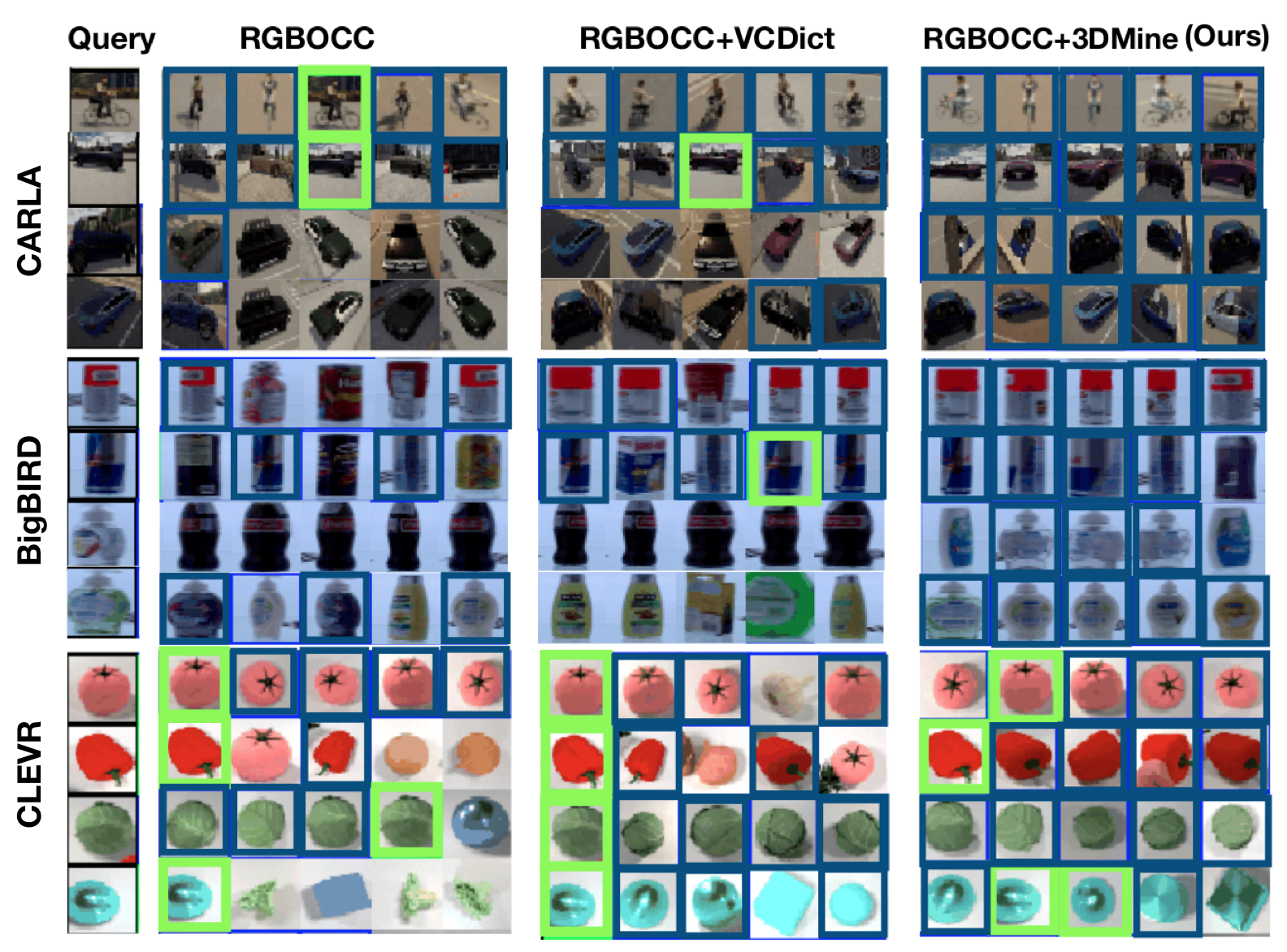}
      \caption{
      \textbf{3D object retrieval results} obtained by retrieving image patch using features learned from different feature learning methods, including
      rgbocc, rgbocc+vcdict, and rgbocc+3D correspondence mining (3DMine) methods. We visualize the retrieval results on CARLA, BigBIRD, and CLEVR datasets. The green boxes indicate that the retrieved image patches belongs to the same object instance as the query, but is in a different viewpoint. The blue boxes indicate instances with the same ground truth object category labels.
    }
    \label{fig:supp_obj_level_retrieval}
\end{figure*}
 
\subsection{Qualitative results for 3D feature representation learning}


Here, we show the qualitative results for object and patch retrieval using the learned 3D visual feature representations from the proposed cross-scene 3D correspondence mining in Section 3.3(main paper). More implementation details are given in Section \ref{sec:imp_details} and Figure \ref{fig:3dmine} of this Appendix.

\paragraph{\normalfont \textbf{Object Level Retrieval.}}

Figure \ref{fig:supp_obj_level_retrieval} shows the qualitative results for object level retrieval.
 Here, we compare the object retrieval results on object-centric (cropped and resized) 3D features maps which are learned from the proposed method (rgbocc + 3D correspondence mining) and 2 other baselines: rgbocc and rgbocc+vcdict, which are detailed in Section 4.3(main paper). 
We show the results on 3 datasets: CARLA, BigBIRD, and CLEVR. For each query image, shown in the first column, we show the top 5 retrievals for the three methods mentioned above. The green box signifies that the retrieved image belongs to the same object category as the query, but is in a different viewpoint of the same scene. Blue box depicts retrieval of the same object category from a completely different scene. As can be seen, our method (rgbocc+3D mining) gives much more accurate retrievals (more number of blue and green boxes) compared to the other two baselines across all datasets. We show the object level retrieval results for this method on Replica dataset in Figure~\ref{fig:supp_habitat_obj_ret}.

\begin{figure*}[h!]
    \centering
    \includegraphics[width=0.95\textwidth]{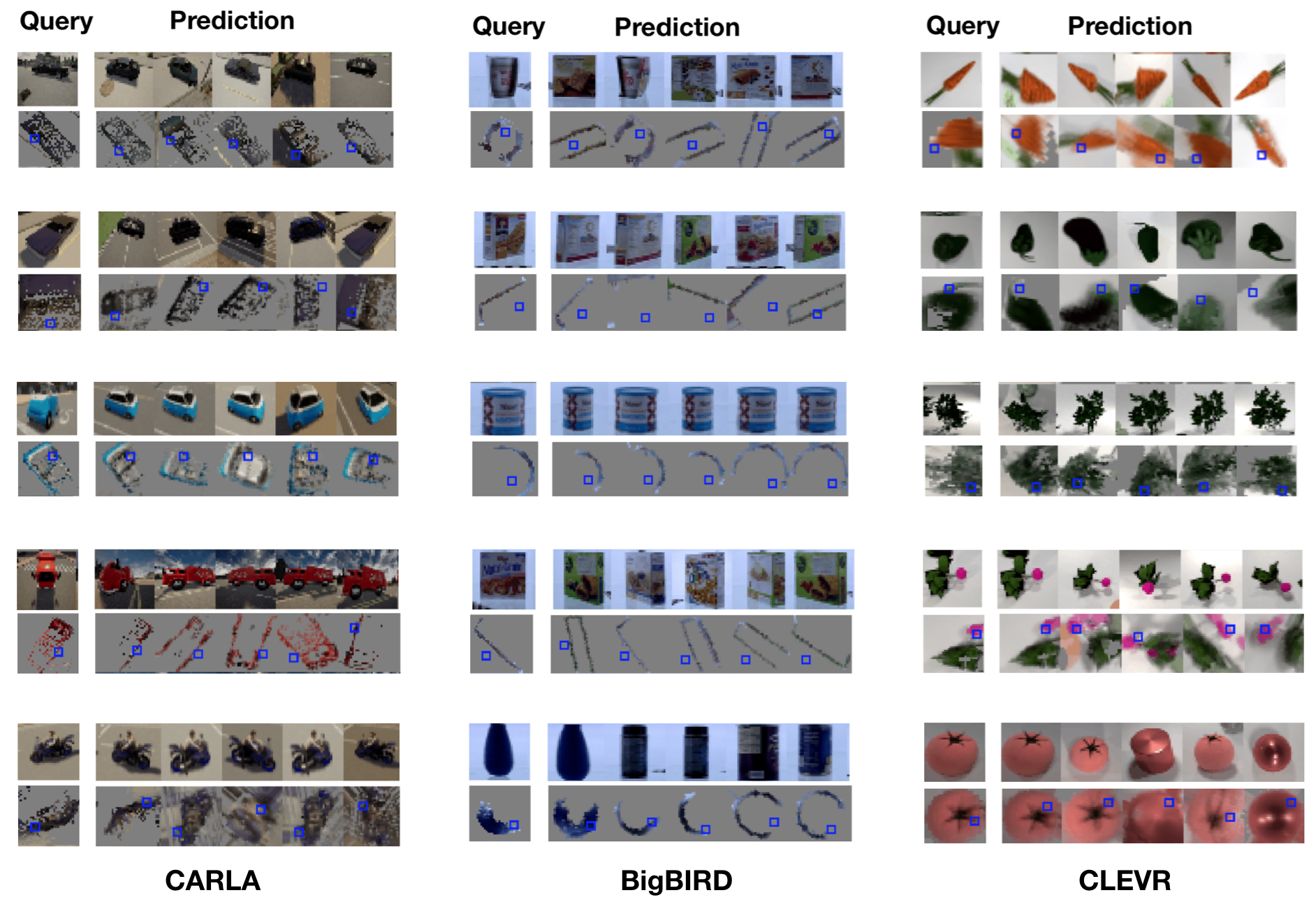}
      \caption{
      \textbf{Patch based 3D object retrieval results} on CARLA, BigBIRD, and CLEVR datasets. For each query-prediction row pair, the first row shows the input RGB images and the second row shows bird's eye view projection of the RGB-D point cloud. The blue patches in the bird's eye view visualizations (2nd row) show the 2D projection of the query/retrieved 3D patch.
    }
    \label{fig:supp_patch_level_retrieval}
    \vspace{-0.5cm}
\end{figure*}

\paragraph{\normalfont \textbf{Patch Based Retrieval.}}
Figure \ref{fig:supp_patch_level_retrieval} shows the 3D object patch retrieval results using the learnt 3D features from the proposed cross-scene 3D correspondence mining technique.   
We visualize the top 5 object part retrievals given a query object patch and a pool of target objects. For each query image, we first unproject it in the 3D space, detect objects in the scene, and randomly select a 3D patch on one of the objects. 
The first column for each dataset represents the query and the next 5 columns show the corresponding top 5 retrieved patches. For each query-prediction row pair, the first row shows the input RGB images and the second row shows bird's eye view of the same RGB images unprojected in 3D space. The blue patches in the bird's eye view visualizations (2nd row) show the 2D projection of the query/retrieved 3D patch. We additionally show patch based retrieval results on Replica dataset in Figure~\ref{fig:supp_habitat_patch_ret}.
We show the top 5 retrieved 3D patches that best matched the corresponding query patch using verification from surrounding voxels technique described in Figure \ref{fig:3dmine} (b).
As can be seen, patch based retrievals seem meaningful when surrounding context is given importance.

\begin{figure*}[h!]
    \centering
    \includegraphics[width=0.85\textwidth]{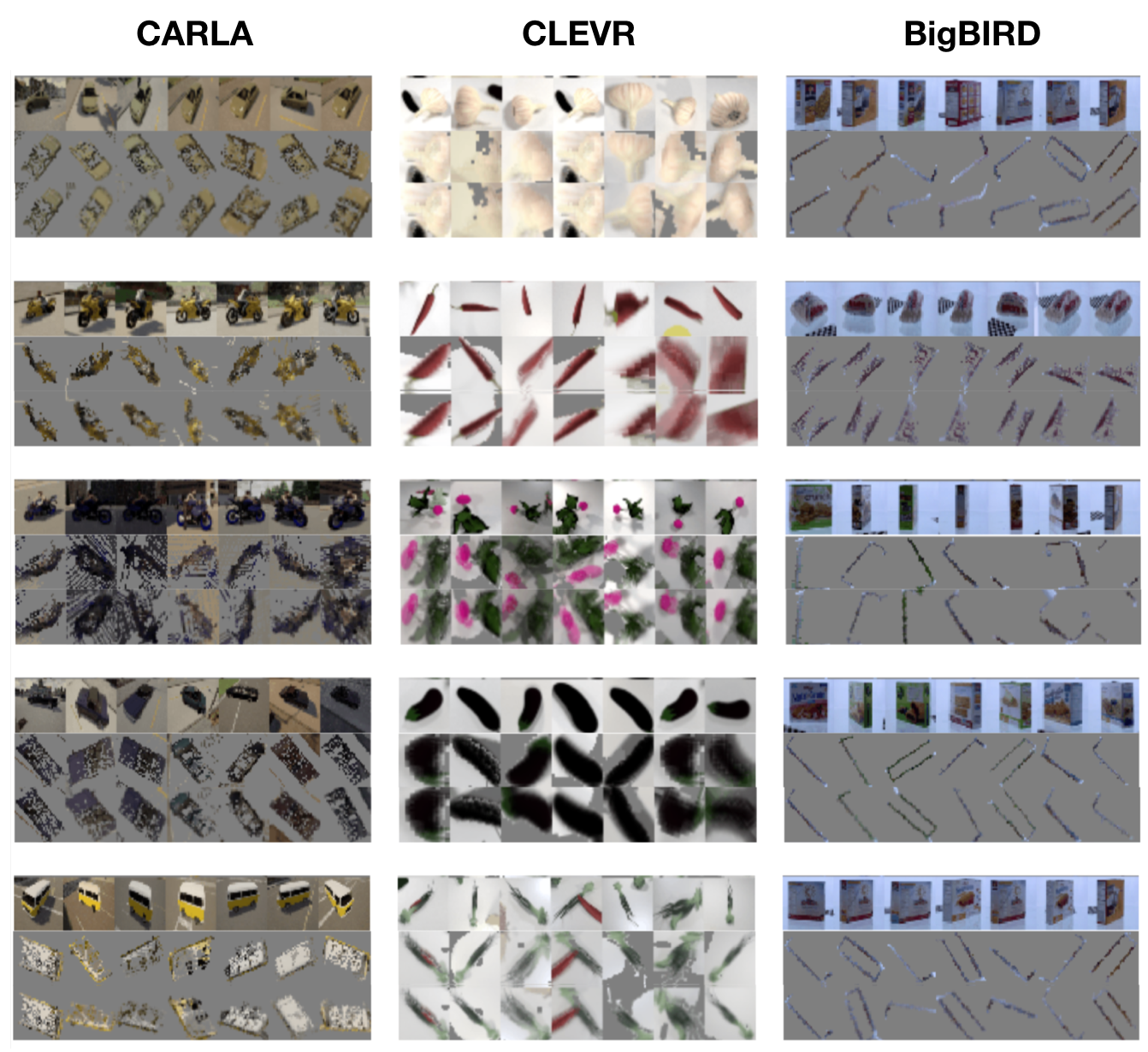}
      \caption{
      \textbf{Rotational alignment results} showing relative pose estimation between two randomly posed RGBs of the same object category. For each of the $3 \times 7$ grids, the first row shows 7 input RGB images of the same object category in different poses. The second row shows the projection of the RGB-D point cloud in a bird’s eye view. The last row shows the projection of the same RGB-D point but warped to the pose that best matches with the object in the first. Results are shown on CARLA, BigBIRD and CLEVR datasets. 
    }
    \label{fig:supp_rotational_align}
\end{figure*}

\paragraph{\normalfont \textbf{Rotation Matching.}}
Finding the rotation transformation between two randomly posed RGB images is a crucial step for our model. As mentioned in Section 3.2(main paper), to do pose-equivariant quantization, we need to first align the input object 3D feature tensors with an object prototype. The quality of our quantization relies on the quality of the features that will yield the correct rotation alignment. 
We show the qualitative performance of such rotation assignment on CARLA, BigBIRD and CLEVR datasets in Figure \ref{fig:supp_rotational_align}.
For each of those $3\times7$ grids, the first row shows the input RGB images of the same object category in different poses, the second row shows the bird's eye view of the same RGBs unprojected in 3D space, and the third row shows the bird's eye view of the same unprojected RGBs but warped to the pose that best matches with the object in the first. We conduct this matching on top of our 3D feature space by doing a rotation aware search. As shown in the visualizations, our model can warp the objects in different orientations to an orientation in the vicinity of the pose of the target object.

\begin{figure*}[h!]
    \centering
    \includegraphics[width=0.6\textwidth]{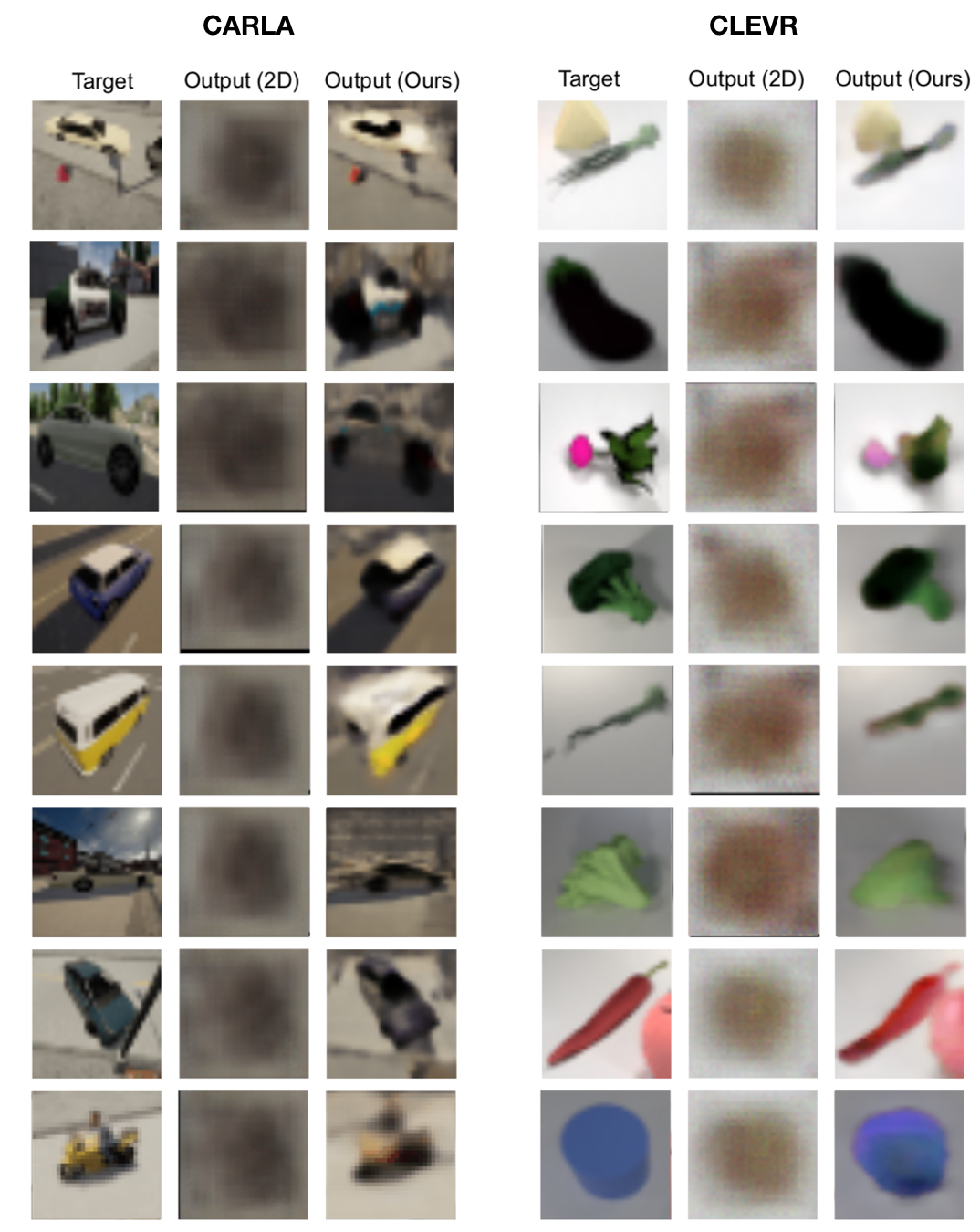}
      \caption{
    Prototype reconstruction results using the learned 3D prototypes from our model and their comparison with 2D prototypes.
    }
    \label{fig:supp_prototype_recon}
\end{figure*}

\subsection{Qualitative results for scene reconstruction using learned 3D object prototypes}

\begin{figure*}[h!]
    \centering
    \includegraphics[width=.9\textwidth]{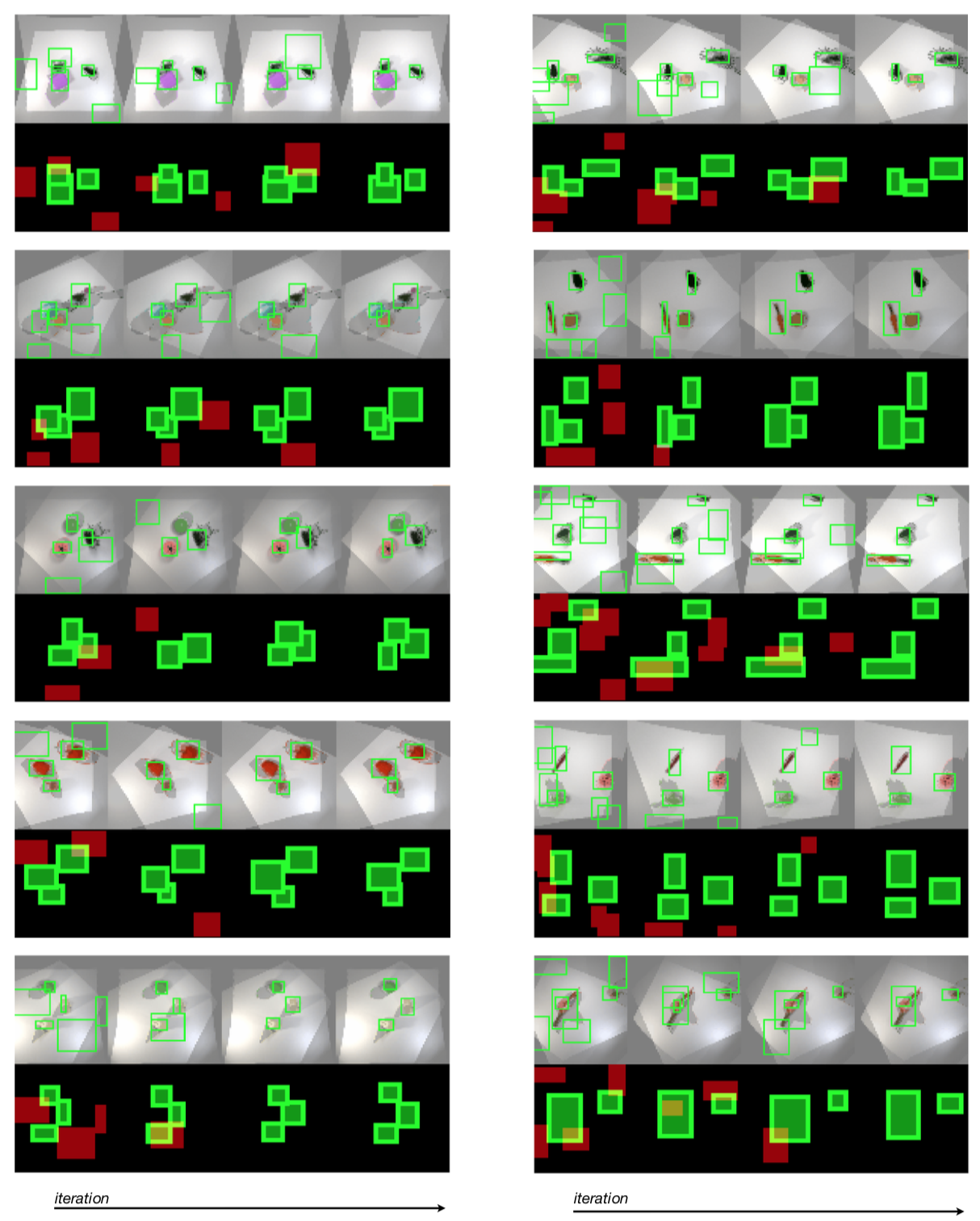}
      \caption{
      The outputs from the proposed iterative detector improvement for 4 iterations. 
    }
    \label{fig:supp_self_improving_det}
\end{figure*}

Learning 3D prototypes is a fundamental part of our pipeline as it helps us in inferring object associations and poses across different scenes. 
In this section, we compare the RGB neural reconstruction of a scene after replacing the objects in the scene with the prototypes learned using our model and the 2.5DQ-Nets baseline on both CARLA and CLEVR datasets. 
In Figure \ref{fig:supp_prototype_recon}, we show the neural scene reconstructions after the object-prototype replacement. For our model we show the 2D neural render of the scene at a different camera view than the input view, whereas for the 2.5DQ-Nets baseline we reconstruct the image at the same camera view. 

For each dataset, the first column represents the ground truth RGB render of the scene. Third column represents the target view  neural render of the scene using our learned 3D prototypes. This reconstruction is obtained by lifting the 2.5D input to 3D feature space, extracting the object from this feature space, finding the best matching prototype, warping the prototype to the pose of the input object, replacing the object features with the warped 3D prototype features, and finally performing RGB view prediction to the target view with these 3D features. The second column shows the reconstruction results when we follow the same procedure as before but use 2D prototypes and 2D rotation check instead of 3D. The 2D prototypes, because of their inability to be 3D rotation-equivariant while quantization, end up learning the mean representation of objects in different poses, which appears as a circular blur. The 3D prototypes, on the other hand, give sharp reconstructions because the objects in different poses are mapped to the same canonical pose. 
Results on the Replica dataset are shown in Figure \ref{fig:supp_habitat_proto_recons}.

\subsection{Qualitative results for the self-improving object detector}


\noindent
As mentioned in Section 3.4(main paper), our model can use the learned 3D object prototypes to self-improve its object detection. The object detector will propose several 3D bounding box proposals. The model will then self label some of these proposals as good proposals if the content inside the proposal can be well-explained by the learned 3D object prototypes, and will label it as bad proposals if the content inside the proposal is not salient and has low 3D center-surround score.
The object detector then uses these pseudo labels to refine its weights to achieve better detection. We visualize the detections made by our self-improving detector on CLEVR dataset over 4 iterations in Figure \ref{fig:supp_self_improving_det}. The first row of the figure represents the bounding boxes predicted by our detector at each iteration. The second row shows the self annotated labels generated using the prototype distance and center-surround score for each bounding box at each iteration. The negative boxes which are to be pruned are shown in red, and the ones to be kept are shown in green. As can be seen, our model can propose accurate positive and negative labels to the proposed 3D boxes. Although the object detector performs poorly in the first iteration, the quality of the object proposals made by our object detector can improve over iterations and produce accurate results after 4 iterations.
\begin{figure*}[h!]
    \centering
    \includegraphics[width=.90\textwidth]{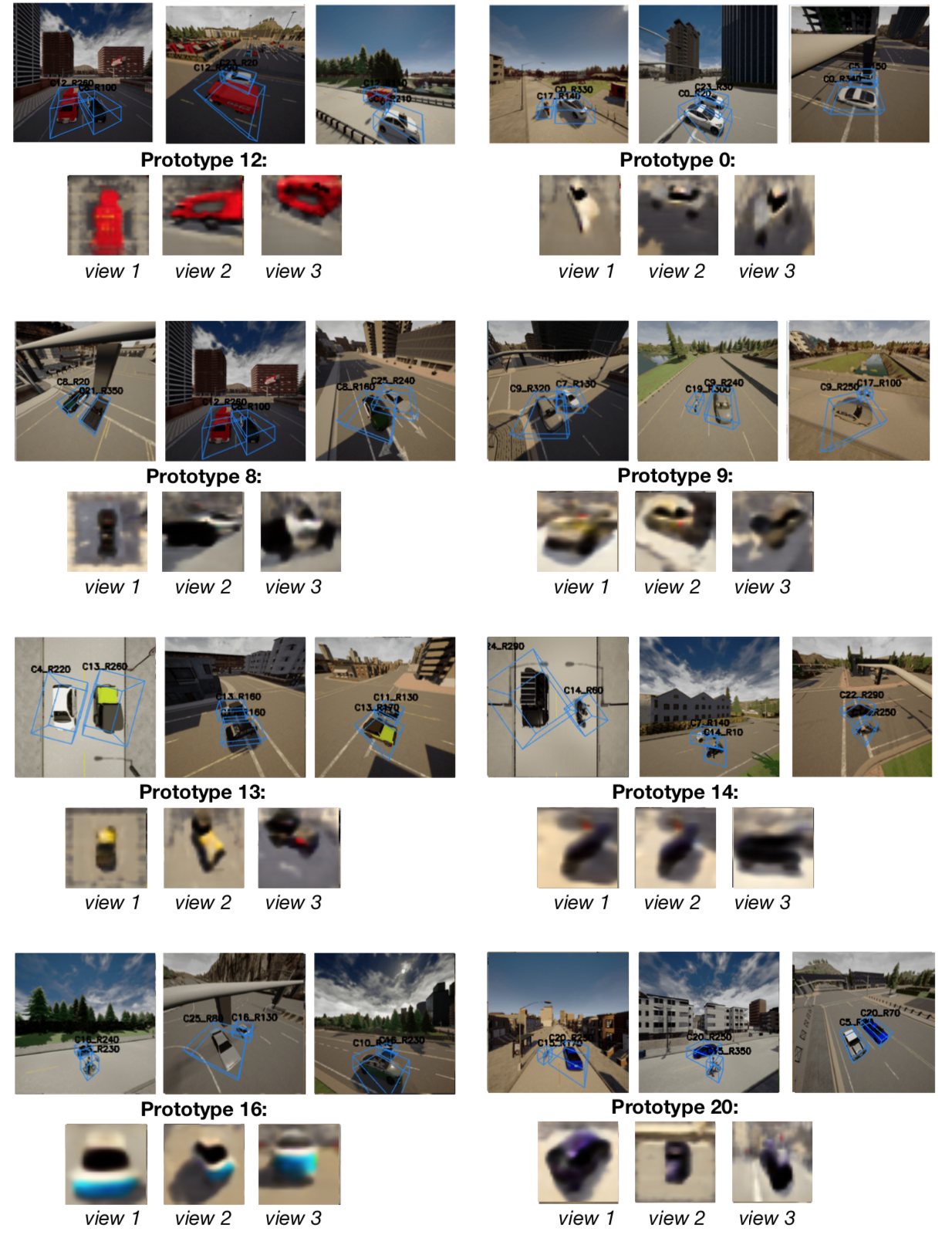}
      \caption{
      Scene parsing results for CARLA dataset.
    }
    \label{fig:supp_carla_scene_parse}
\end{figure*}
\subsection{Qualitative results for scene parsing}

\begin{figure*}[h!]
    \centering
    \includegraphics[width=0.98\textwidth]{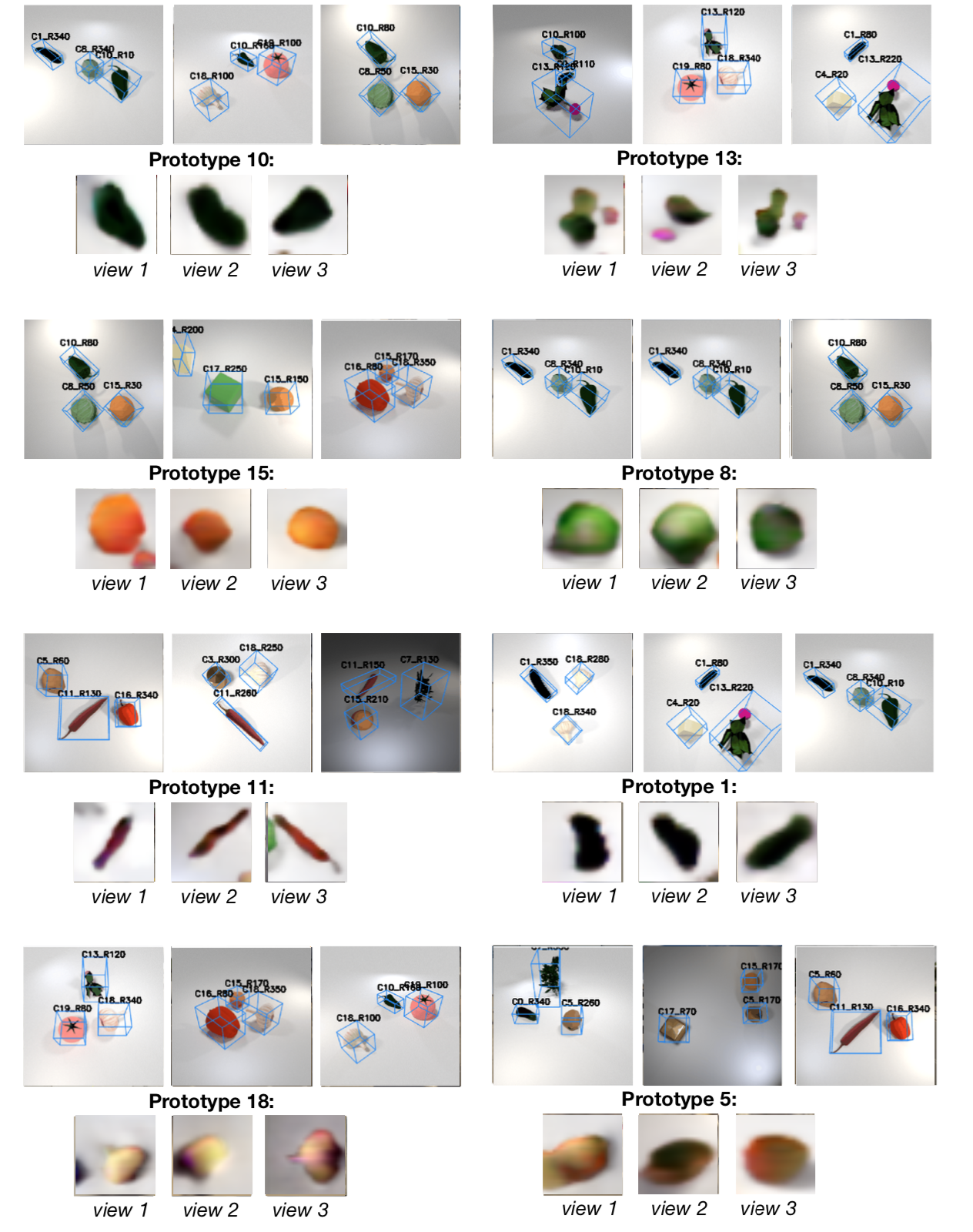}
      \caption{
      Scene parsing results for CLEVR dataset.
    }
    \label{fig:supp_clevr_scene_parse}
\end{figure*}

In this section, we show that using our learned prototypes and  self improved 3D object detector, we are capable of parsing a scene at test time.
Figure~\ref{fig:supp_carla_scene_parse}, Figure~\ref{fig:supp_clevr_scene_parse}, and Figure~\ref{fig:supp_habitat_scene_parse} show the scene parsing results of our 3DQ-Nets on the CARLA, CLEVR, and Replica datasets, respectively. Our model is able to learn 3D scene parsing without any 3D supervision. As explained in the paper and shown in the figures, 3DQ-Nets can learn to infer 3D scene parsings from the input RGB-D images.

The top row in each visualization in both the figures shows the parsing of a scene by drawing the inferred bounding boxes, along with a text on top of each box mentioning the Prototype Number(C) and Rotation angle(R) in degrees. For example, an object associated to Prototype 5 with a Relative Rotation of $150\degree$ with respect to its associated cluster 5 prototype is represented as 
\textit{C5\_R150}.
In the next row for each visualization we also show 3 different camera view neural renders of the prototypes, while also mentioning their respective prototype numbers.
These neural renders of prototypes are generated by placing the prototype in randomly selected backgrounds.
Note that the rotation angles shown in the results are relative to the  pose of the reference camera used for recording the scene. Since we randomize the orientation of the reference camera for each scene, it is possible that two objects in seemingly similar poses have different relative rotation angles.


\begin{figure*}[h!]
    \centering
    \includegraphics[width=0.5\textwidth]{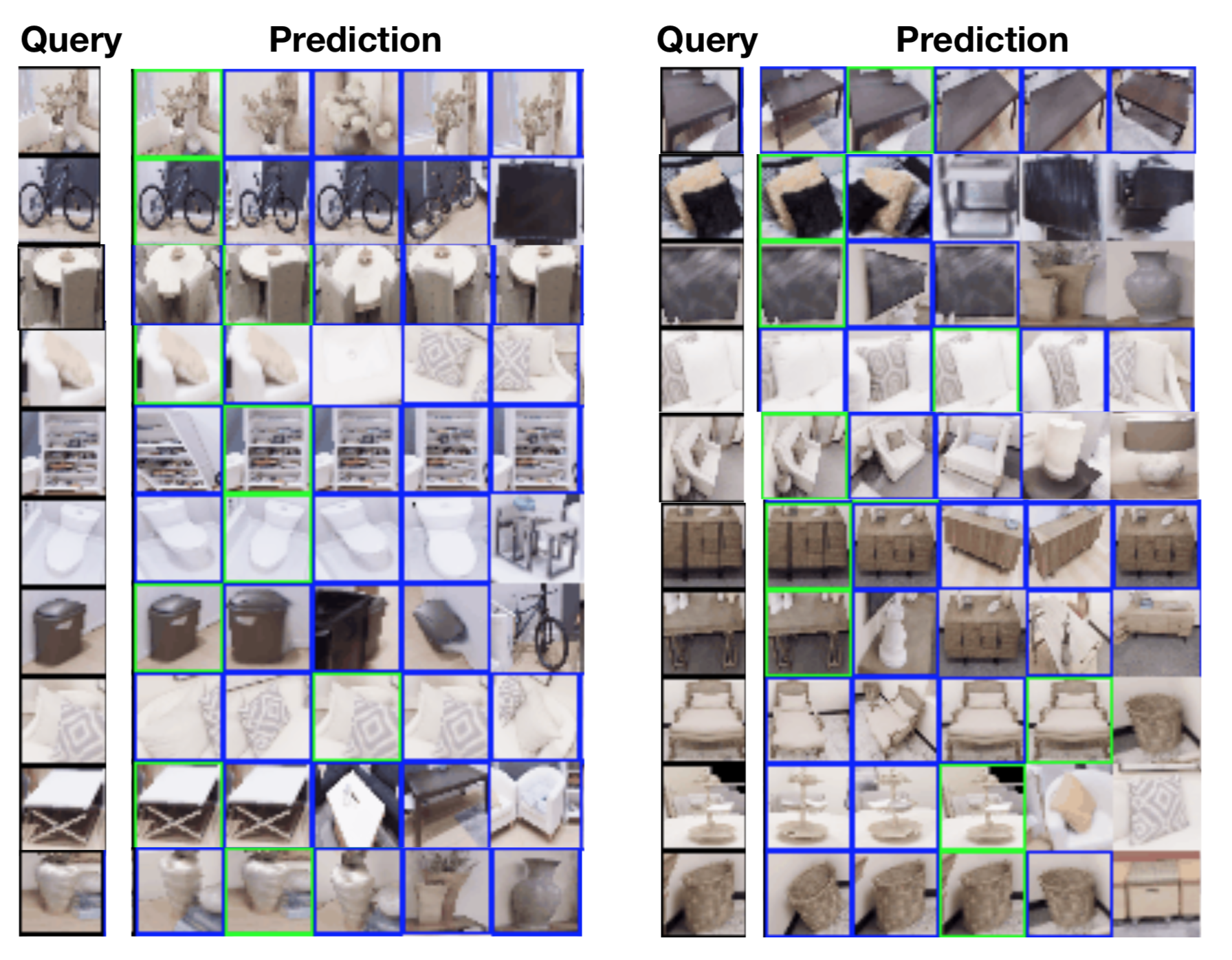}
      \caption{
      3D object retrieval results obtained by rgbocc+3D correspondence mining on Replica dataset.
    }
    \label{fig:supp_habitat_obj_ret}
\end{figure*}

\begin{figure*}[h!]
    \centering
    \includegraphics[width=0.5\textwidth]{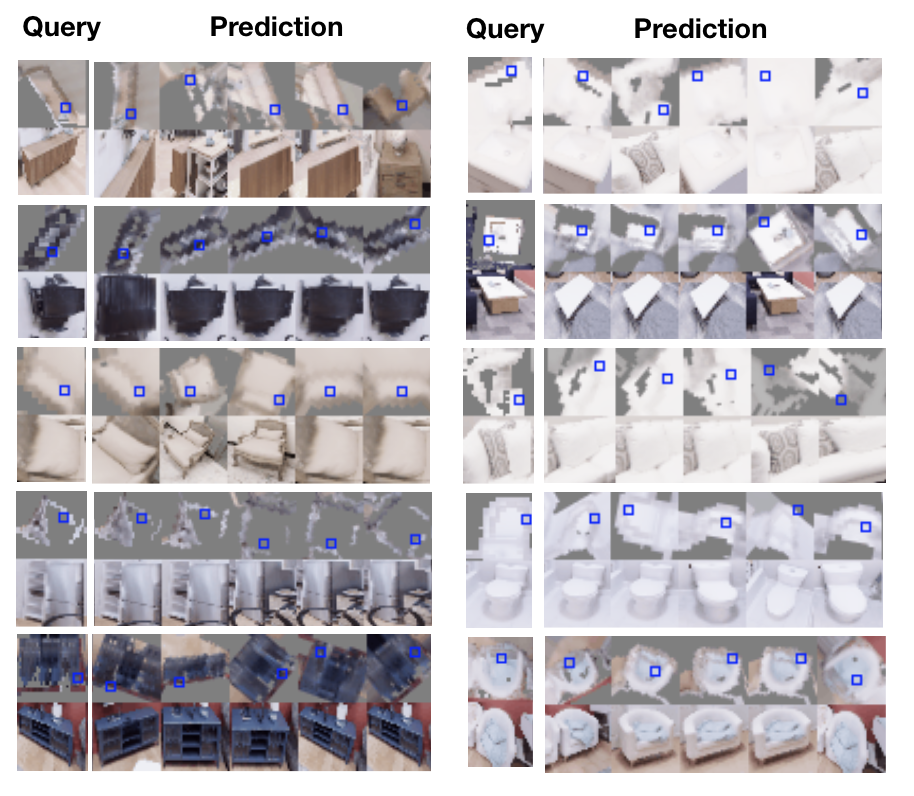}
      \caption{
      Patch based 3D object retrieval results on Replica dataset.
    }
    \label{fig:supp_habitat_patch_ret}
\end{figure*}

\begin{figure*}[h!]
    \centering
    \includegraphics[width=0.3\textwidth]{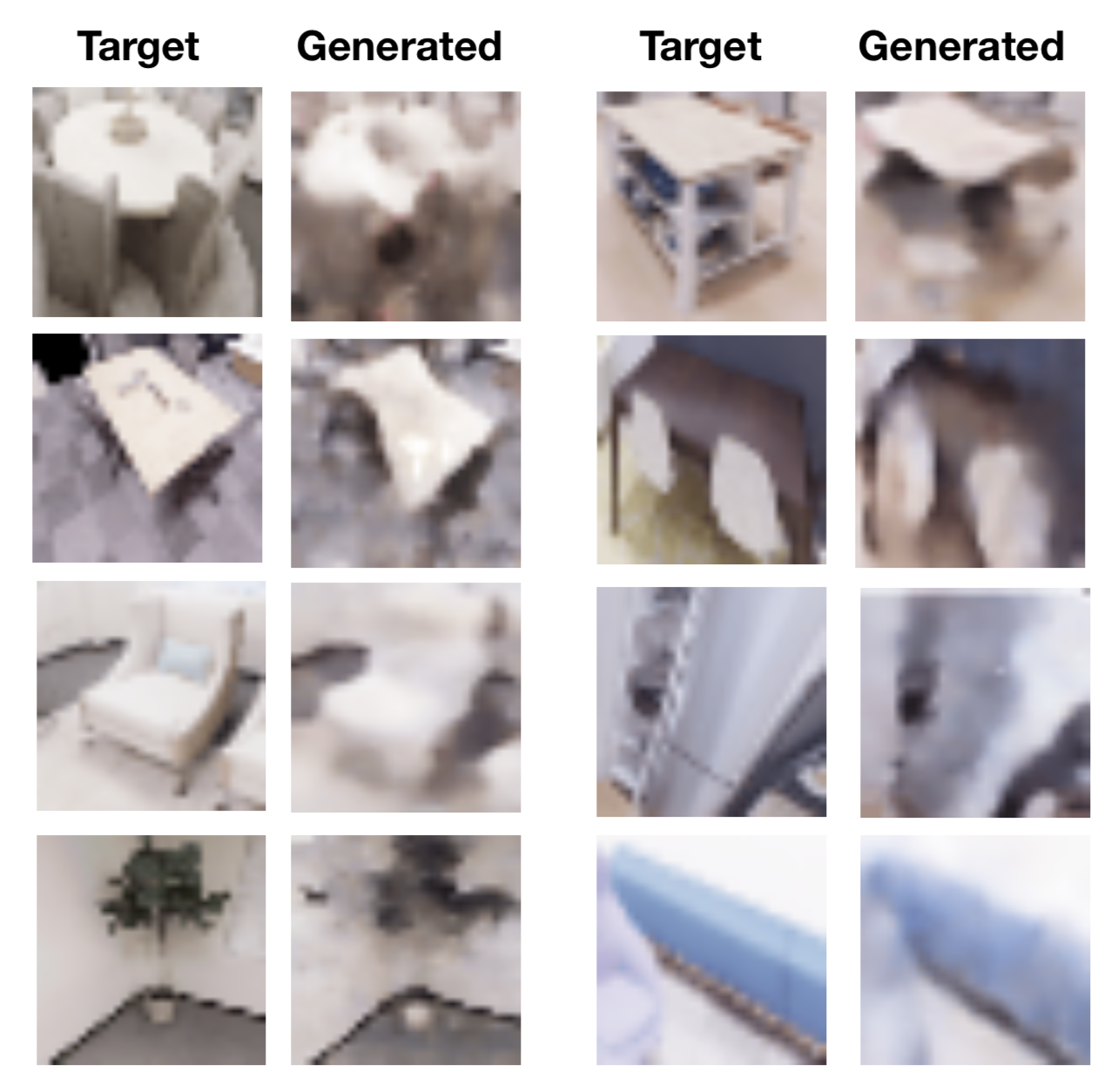}
      \caption{
      Prototype reconstruction for 3D prototypes learned by our model on Replica dataset.
    }
    \label{fig:supp_habitat_proto_recons}
\end{figure*}

\begin{figure*}[h!]
    \centering
    \includegraphics[width=0.98\textwidth]{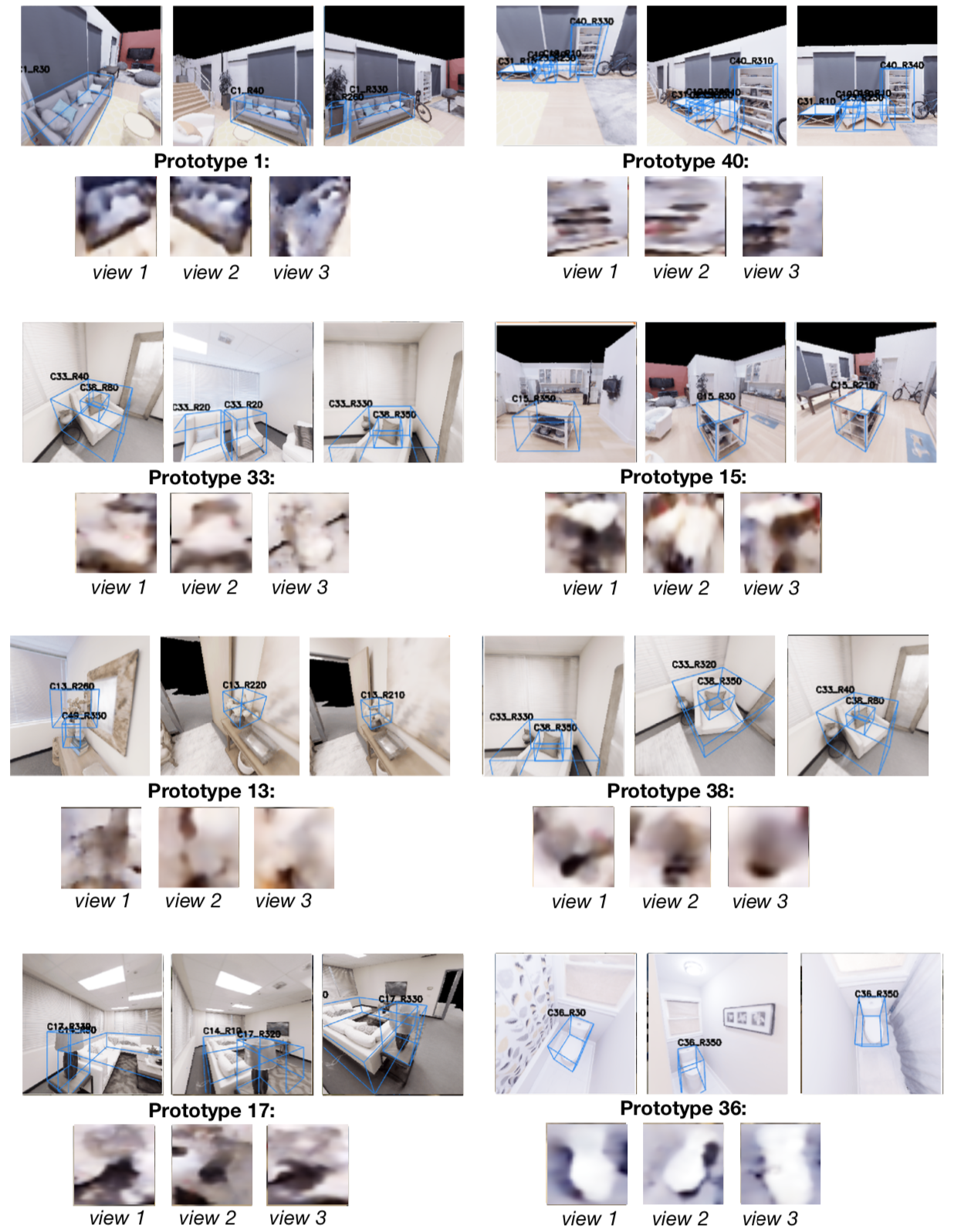}
      \caption{
      Scene parsing results for Replica dataset.
    }
    \label{fig:supp_habitat_scene_parse}
\end{figure*}


%
%